\documentclass[11pt]{article}

\usepackage{acl}

\usepackage{times}
\usepackage{latexsym}
\usepackage{amsmath}
\usepackage{amsthm}
\theoremstyle{definition}
\newtheorem{definition}{Definition}

\usepackage[T1]{fontenc}

\usepackage[utf8]{inputenc}

\usepackage{microtype}

\usepackage{inconsolata}

\usepackage{graphicx}

\usepackage{booktabs}
\usepackage{pifont}
\usepackage{xcolor}
\usepackage{colortbl}
\usepackage{array}
\usepackage{makecell}
\usepackage{subcaption}
\usepackage{multirow}

\newcommand{\cmark}{\textcolor{teal}{\ding{51}}}
\newcommand{\xmark}{\textcolor{gray}{\ding{55}}}

\newcolumntype{L}[1]{>{\raggedright\arraybackslash}m{#1}}
\newcolumntype{C}[1]{>{\centering\arraybackslash}m{#1}}

\title{Intrinsic Attention as Navigator: A Non-Generative Retrieval Method for Multi-Hop Queries}

\author{
  \textbf{Linwei Song\textsuperscript{a}},
  \textbf{Weijie Zhang\textsuperscript{a}},
  \textbf{WeiDi Chen\textsuperscript{a}},
  \textbf{Huixia Lai\textsuperscript{*,a}},
  \textbf{Shi Zhang\textsuperscript{*,a,b}},
  \textbf{Ruliang Xiao\textsuperscript{a,b}}\\
  \textnormal{[a] College of Computer and Cyber Security, Fujian Normal University, Fuzhou 350117, China}\\
  \textnormal{[b] Digit Fujian Internet-of-Things Laboratory of}\\
  \textnormal{Environmental Monitoring, Fuzhou 350117, China}\\
  \normalfont\small{
    \textbf{*Corresponding authors:}
    \href{mailto:lhx@fjnu.edu.cn}{\texttt{lhx@fjnu.edu.cn}}, \href{mailto:shi@fjnu.edu.cn}{\texttt{shi@fjnu.edu.cn}}
  }
}
\begin{document}
\maketitle

\begin{abstract}
For multi-hop queries in retrieval-augmented generation (RAG), it is crucial to accurately obtain evidence pieces. In this study, we propose a training-free method for multi-hop query retrieval, named Intrinsic Navigating-based Non-Generative Retrieval (IN-Retriever). Instead of generating intermediate reasoning text, the method extracts the bridging clue from the internal attention of the embedding model based on the current context and retrieved candidate documents. During iterative clue extraction and document retrieval, IN-Retriever strips away revealed clues from the embedded query vector and then reranks candidate documents in a targeted manner, thereby steering retrieval toward hidden next-hop evidence. In the experiments, we confirm the feasibility and effectiveness of using intrinsic attention signals to guide external multi-hop retrieval. Experimental results show that our method precisely collects sufficient documents to support multi-hop queries and significantly outperforms sparse and dense retrieval methods.
\end{abstract}

\section{Introduction}
Retrieval-augmented generation (RAG) mitigates the limitations of large language models (LLMs) in knowledge timeliness and domain-specific expertise by retrieving information from external data sources \citep{lewis2020rag,gao2024ragsurvey}. In RAG, retrieving sufficient evidence is crucial for LLMs to generate the final answer. As to multi-hop question answering (QA), RAG systems need to progressively gather evidence across multiple documents. Existing single-pass retrieval methods rely directly on the user query, which can not guarantee the retrieved documents jointly forming a complete evidence chain \citep{xiong2021multihopdense}.

To address the above issue, some studies explicitly generate intermediate reasoning steps and then design retrieval paths \citep{trivedi2023interleaving, yao2023react,asai2024selfrag}. In sequential generation, LLMs may over-commit to early mistakes, a phenomenon known as hallucination snowballing \citep{zhang2023sirenssong}. Other studies improve dense dual-encoder retrieval through supervised training and hard-negative selection \citep{karpukhin2020dpr,xiong2021annncl}. Single-vector dual encoders represent each query and document with one fixed-dimensional vector and score relevance with a single dot product \citep{ni2022gtr}. ColBERTv2 retains token-level representations and applies late interaction for finer-grained query-document matching \citep{santhanam2022colbertv2}.

Recently, \citet{yu2023factualrecall} identified individual attention heads in Pythia and GPT-2 that promote either memorized or in-context answers in a counterfactual world-capitals task. By rescaling the heads' value vectors, they changed the relative likelihood of the two answers at runtime. \citet{gould2024successorheads} identified successor heads that increment tokens in natural ordinal sequences and found a weak form of universality across model architectures and scales. These results indicate that individual attention heads can play predictable roles in model behavior. We therefore examine whether attention signals in an embedding model can identify bridging clues and develop Intrinsic Navigating-based Non-Generative Retrieval (IN-Retriever), a training-free method for multi-hop queries. This method guides document retrieval by leveraging the internal attention signals of embedding models, without explicitly generating reasoning text. Specifically, it first activates specific attention heads in the embedding model through instruction-guided prompting, and then extracts bridging clues from the current context based on the model's internal attention to retrieve candidate documents for the next hop. It subsequently strips away revealed clues from the embedded query vector and reranks the candidate documents, thereby steering retrieval toward hidden next-hop evidence.

Overall, the main contributions of this paper are summarized as follows.

(1) We propose a non-generative retrieval method for multi-hop queries, named the Intrinsic Navigating-based Non-Generative Retrieval method (IN-Retriever). The method exploits the internal attention heads of embedding models to discover bridging clues for multi-hop queries.

(2) We verify the implicit relationship between the internal attention of embedding models and bridging clues, and propose an instruction-driven method for locating elite attention heads for obtaining bridging clues with no explicit reasoning text.

(3) We design a residual-guided reranking module, which gradually removes the semantic component in the query vector that has already been expressed by the current clue, and reranks candidate documents using the remaining information, thereby steering retrieval toward next-hop evidence.

(4) Experiments show that IN-Retriever outperforms the feasibility and effectiveness in multi-hop queries, demonstrating a feasible way to leverage the internal attention signals of large models to locate bridging clues.

\section{Methodology}
\label{section:methodology}

\textbf{Problem Definition:}
Given a query \(Q\) and a large-scale unstructured corpus \(C\), the aim of retrieval is to extract and construct an evidence document, such that these documents can jointly support the reasoning toward the answer \(A\) for query \(Q\).

\subsection{Framework overview}
\begin{figure*}[t!]
\vspace{-0.2cm}
  \centering
  \includegraphics[width=1.0\textwidth]{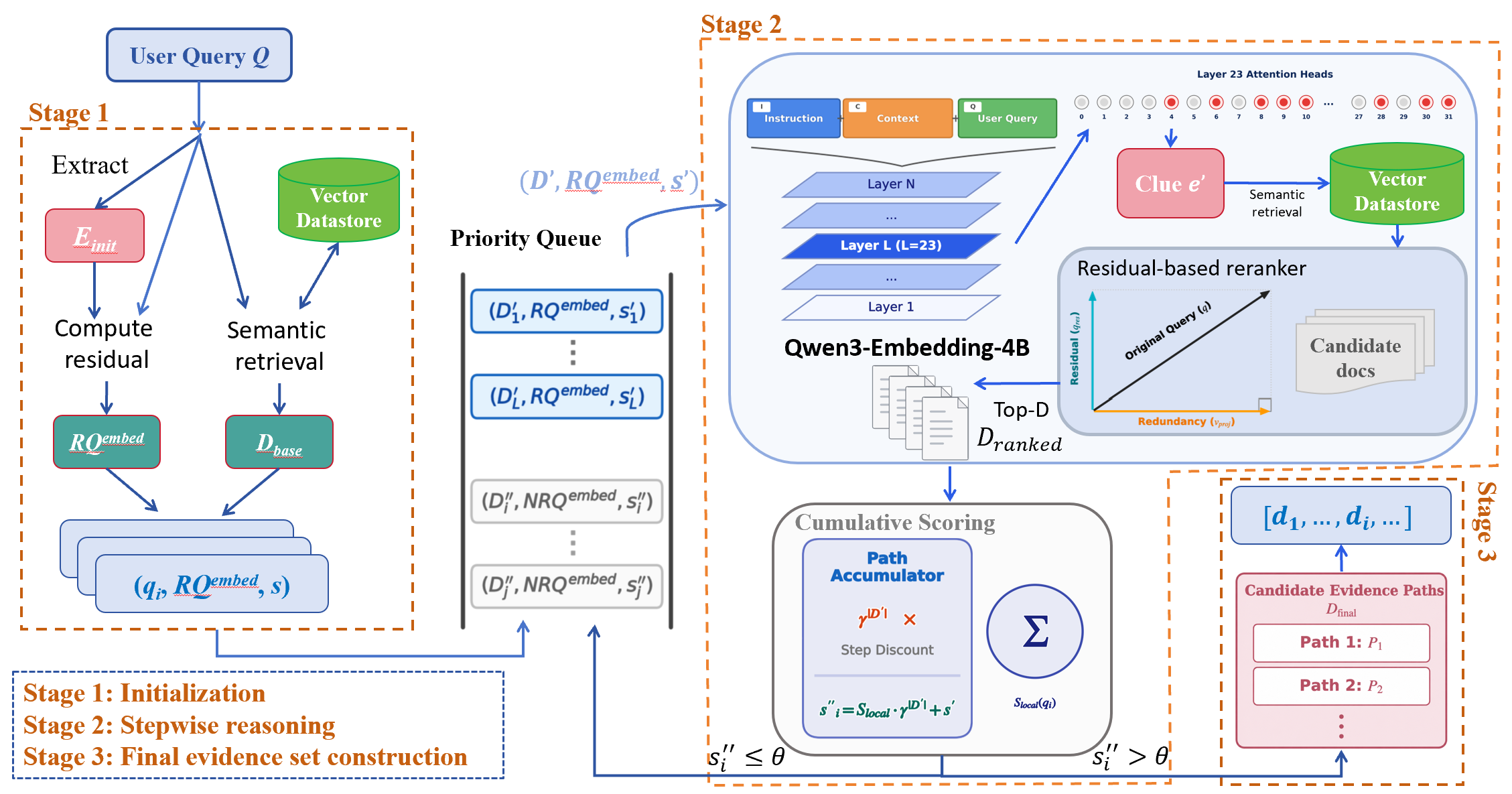}
  \caption{Framework of the Intrinsic Navigating-based Retrieval Method (IN-Retriever). The overall process is delineated into three stages: initialization, stepwise reasoning, and final evidence set construction. The initialization stage retrieves documents grounded on semantic similarity. The stepwise reasoning stage employs a beam search mechanism to progressively expand the document evidence chains. Ultimately, the trajectory attaining the highest cumulative score is selected to yield the final evidence set.}
  \label{fig:arch}
\vspace{-0.2cm}
\end{figure*}

The structure of IN-Retriever is illustrated in Figure \ref{fig:arch}. The retrieval process can be divided into three stages: initialization, stepwise reasoning, and final evidence set construction.

\textbf{Initialization stage}. First, the potential named entities and noun phrases contained in the query $Q$ are extracted as an initial entity set $E_{init}$. Then, the most similar $J$ documents are retrieved from the Vector Datastore (VD) based on semantic retrieval, forming $D_{base}$. Subsequently, the retrieval-intent residual $RQ^{\text{embed}}$ is computed based on the user query embedding $Q^{\text{embed}}$ and $E_{init}$, where the residual computation is defined in Section \ref{subsection:residual}. Finally, $RQ^{\text{embed}}$, the document $q_i \in D_{base}$, and score $s$ are combined into $J$ triples $([q_i], RQ^{\text{embed}}, s)$ and inserted into the priority queue $PQ$, where $s$ denotes the cumulative score of the evidence chain $[q_i]$ and serves as the priority metric in the priority queue, and the scoring mechanism is defined in Section \ref{subsection:scoring}.

\textbf{Stepwise reasoning stage}. Elements are repeatedly popped from $PQ$ and processed. For each element $(D', RQ^{\text{embed}}, s')$, the process consists of three steps as follows.

(1) The instruction $I$, the last document $q'$ in the document list $D'$, and the original query $Q$ are fed into the embedding model. Next, a bridging clue $e'$ is extracted from $q'$ using the instruction-driven bridging clue probing method (Section \ref{subsection:clueProbing}). Subsequently, $e'$ is used to retrieve a candidate document set $D_{cand}$ from the VD.

(2) Based on $RQ^{\text{embed}}$, the semantic residual-based reranking method (Section \ref{subsection:reranking}) is applied to $D_{cand}$ to obtain the top-\(D\) documents $\{q_1, \ldots, q_D\}$.

(3) The $D$ documents are individually appended to $D'$, forming $D$ document lists
$\{D''_1, \ldots, D''_D\}$. Next, the retrieval-intent residual ${IRQ}^{\text{embed}}$ is calculated based on $RQ^{\text{embed}}$ and $e'$. Based on $s'$, the cumulative score $s''_i$ of each new evidence chain $D''_i$ is calculated. If $s''_i$ is greater than the threshold $\theta$, the evidence chain $D''_i$ is added to $D_{final}$ as a candidate evidence path; otherwise, a new triple $(D''_i, IRQ^{\text{embed}}, s''_i)$ is pushed into $PQ$.

The stepwise reasoning expands the reasoning evidence chain. To avoid unlimited expansion, IN-Retriever also limits the expansion path length. 

\textbf{Final evidence set construction stage}. Evidence paths are selected from $D_{final}$ in descending order of score. The documents are merged as final retrieval results.

IN-Retriever adopts the beam search mechanism \citep{zhao2021beamdr}, which allows it to maintain multiple potential reasoning trajectories (trajectory hypotheses) simultaneously, thereby enabling the model to tolerate local uncertainty. 

In the remainder of this section, we will introduce the instruction-driven bridging clue probing method (Section \ref{subsection:clueProbing}), retrieval-intent residual computation (Section \ref{subsection:residual}), semantic residual-based reranking (Section \ref{subsection:reranking}), and the cumulative scoring mechanism for evidence chains (Section \ref{subsection:scoring}) in detail.

\subsection{Instruction-Driven Bridging Clue Probing}
\label{subsection:clueProbing}
The instruction-driven bridging clue probing method aims to identify bridging clues supporting next-hop expansion. It consists of two steps: task-oriented attention head selection and bridging clue extraction. Among them, the first step requires only a one-time calibration for a specific embedding model to identify the attention heads. Once the calibration and selection are completed, no further adjustment is required during the subsequent inference process. 

\subsubsection{Task-Oriented Attention Head Selection} %
To steer the model's internal attention toward clues relevant to the next hop, we propose a task-oriented method for selecting key attention heads, termed \emph{elite attention heads}. Specifically, task orientation is implemented through instruction guidance by prepending a task-related instruction $I$ to the input, together with the context $C$ and the user query $Q$. The instruction is defined as:
\vspace{-0.1cm}
\begin{quote}
\small
\textit{Identify the key evidence in the context that helps find the next context to finally answer the question. Focus specifically on the entity type requested in the query.}
\end{quote}
\vspace{-0.1cm}

The probing input is organized as a single sequence $[I;C;Q]$, corresponding to
\texttt{[Instruct; Context; Query]}, where the instruction appears once at the beginning.
Clue extraction is restricted to the token span of $C$, while the instruction and
query tokens only guide bridge clue probing and are never decoded.

This instruction encourages the embedding model to focus on local segments in the context that are relevant to the query, thereby activating specific attention heads and providing more stable clue signals for key attention head selection and subsequent bridging clue probing.

For elite attention head selection, we first construct a calibration dataset $\mathcal{D}_{\text{calib}}$ based on the 2WikiMultihopQA (2Wiki) \citep{ho2020multihopdataset}, where each sample contains a context $C$, a query $Q$, and a bridging clue $E$ extracted via a lexical matching strategy. Then, for each attention head, we compute the Bridge Focus Score (BFS), as defined in Formula \ref{eq:bfs}, to measure the degree of attention allocated to the bridging clue. %
\vspace{-0.1cm}
\begin{equation}
	\label{eq:bfs}
	\begin{aligned}
		&\operatorname{BFS}(h,l)
		= \frac{1}{|\mathcal{D}_{\mathrm{calib}}|}
		\sum_{(C,Q,E)\in \mathcal{D}_{\mathrm{calib}}}
		s_{Q,E}^{(h,l)} \\  
		&s_{Q,E}^{(h,l)}
		= \frac{1}{|T_Q|}  %
		\sum_{i\in T_Q}
		\sum_{j\in T_E}
		A^{(h,l)}_{i,j}
	\end{aligned}
\end{equation}
where $T_Q$ and $T_E$ denote the token-index sets of the query and the annotated bridging clue. $A^{(h,l)}_{i,j}$ is the attention weight from position $i$ to position $j$ by the $h$-th head in the $l$-th layer.

Finally, based on the layer-wise bridge focus score (LBFS) as defined in Formula \ref{eq:lbfs}, we select the $L$-th layer as the key attention layer, and then choose the top-\(H\) attention heads with the highest BFS values in that layer as the elite head set $\mathcal{H}_{\text{elite}}$. 
\vspace{-0.1cm}
\begin{equation}
\operatorname{LBFS}(l)
=
\sum_{h \in \operatorname{argTop}_{H}(\operatorname{BFS}(h,l))}
\operatorname{BFS}(h,l)
\label{eq:lbfs}
\end{equation}
where \(H\) denotes the number of selected attention heads. We set \(H=8\).

According to our empirical analysis, \textbf{for the Qwen embedding model, attention heads in $\mathcal{H}_{\text{elite}}=\{10, 30, 8, 28, 31, 6, 4, 9\}$ at layer $L=23$ are set as elite attention heads}. They are the basis for bridging clue extraction. %
All layer and head indices in this paper follow a zero-indexed convention; therefore,
the 32 heads of Qwen3-Embedding-4B are indexed from 0 to 31.

\subsubsection{Bridging Clue Extraction}
Bridging clue extraction aims to extract text fragments that bridge hops. After feeding $[I; C; Q]$ into the embedding model, we first pinpoint local peaks in $\mathcal{H}_{\text{elite}}$, and then extend them into complete semantic units through locally aware boundary expansion. The extraction method seeks to isolate clues with reasoning value from long documents. It consists of the following steps.

(1) Given the query \(Q\) and the document \(C=\{s_1,\ldots,s_i,\ldots\}\), where \(s_i\) is a sentence in \(C\), we define the average attention weight of each sentence \(s_i\) on the elite attention heads according Formula~\ref{eq:sent_weight}, and then retain the top-\(S\) sentences with the highest weights for the subsequent process. In our experiments, we set \(S=3\).
\vspace{-0.1cm}
\begin{equation}
W(s_i)=\frac{1}{|s_i|}\sum_{t\in s_i}\sum_{h\in \mathcal{H}_{\text{elite}}}\operatorname{Att}(t,h)
\label{eq:sent_weight}
\vspace{-0.1cm}
\end{equation}
where $|s_i|$ is the size of tokens in $s_i$, and $\operatorname{Att}(t,h)$ denotes the attention weight assigned to token $t$ by attention head $h$. 

(2) For the selected \(S\) sentences, we accumulate attention weights of each token over the elite attention heads. Let $t_{j,i}$ denote the $i$-th token in sentence $s_j$. Its cumulative attention weight on $\mathcal{H}_{\text{elite}}$ is defined as Formula \ref{eq:token_cw}.
\begin{equation}
\vspace{-0.1cm}
\operatorname{CW}(t_{j,i})=\sum_{h\in \mathcal{H}_{\text{elite}}}\alpha_{h,i}
\label{eq:token_cw}
\end{equation}
where $\alpha_{h,i}$ denotes the attention weight assigned to $t_{n,i}$ by the $h$-th elite head.

(3) For each selected sentence, we select the top-\(A\) token anchors with the highest cumulative attention scores on \(\mathcal{H}_{\text{elite}}\) as peak anchors. In our experiments, \(A=20\).

(4) For each selected sentence, we take the peak anchors as centers, and extend them bidirectionally to absorb adjacent words until a predefined syntactic-semantic blocking token is reached. The clue is then obtained by decoding the extracted tokens. By extracting one clue from each sentence, we finally acquire a total of S clues.

Overall, based on the internal attention weights of the embedding model and the syntactic structure of the text, the probing method provides a schema for conversing diffuse attention signals to discrete reasoning clues at a low computational cost.

\subsection{Retrieval-Intent Residual Computation}
\label{subsection:residual}

\begin{figure}[htbp]
\vspace{-0.2cm}
  \centering
  \includegraphics[width=\columnwidth]{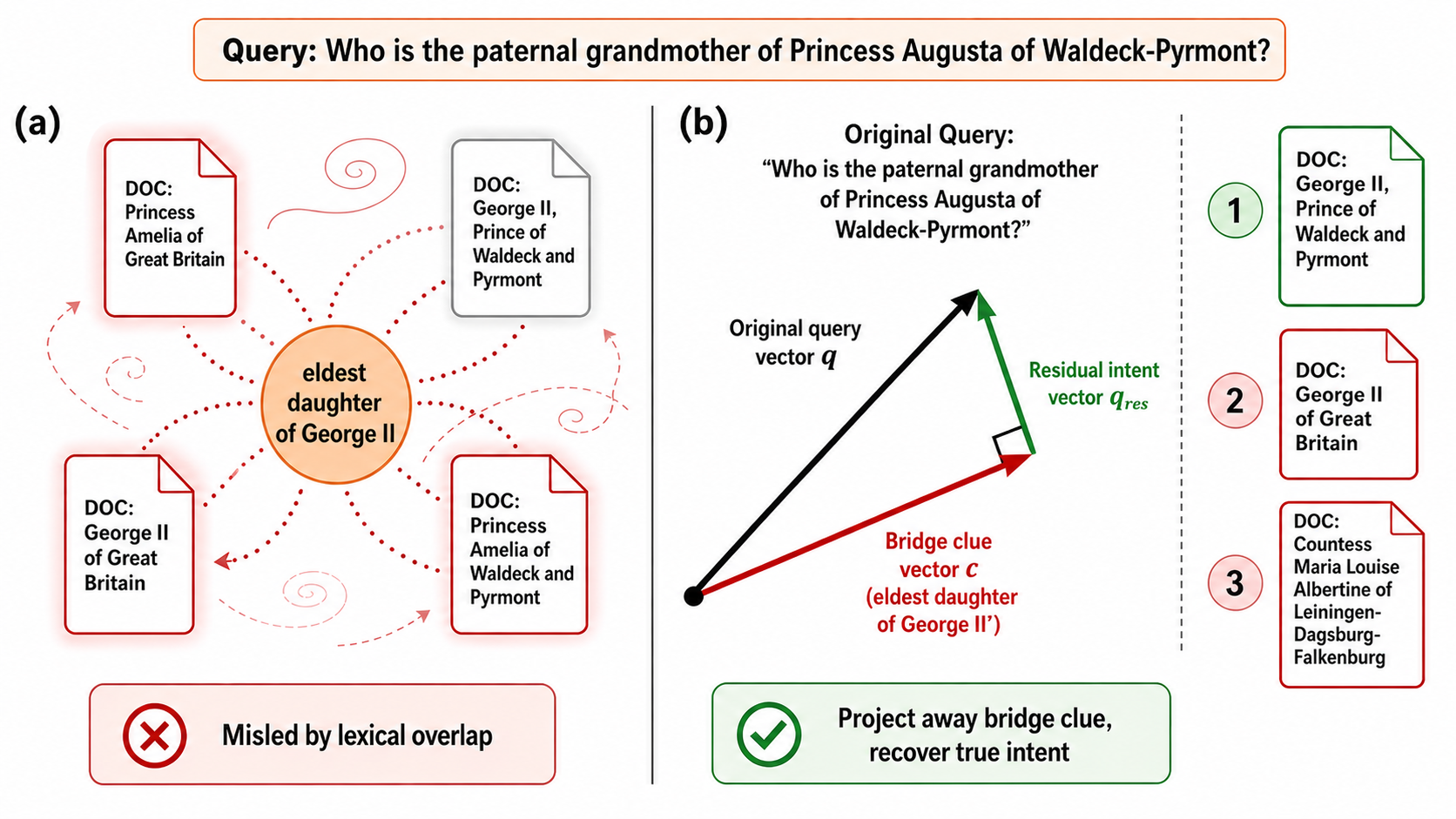}
  \vspace{-0.5cm}
  \caption{Reasoning for the multi-hop query "Who is the paternal grandmother of Princess Augusta of Waldeck-Pyrmont?". (a) Misled by the high lexical frequency of "George II" and "Princess", the dense retriever always recalls documents concerning "King George II of Great Britain" and "Princess Amelia", failing to pinpoint the gold-standard document (George II, Prince of Waldeck). (b) Upon successfully identifying the bridging clue "eldest daughter of George II", the query is projected onto the orthogonal complement of the clue. By isolating the residual intent vector and maximizing its alignment score, the retriever explicitly upweights the target document (George II, Prince of Waldeck and Pyrmont)}
  \label{fig:Retrieval-Intent Residual Computation} 
\vspace{-0.2cm}
\end{figure}

In multi-hop retrieval, relying solely on the original query $Q$ often traps the retriever within a semantic neighborhood highly similar to $Q$. Consequently, even after multiple retrieval rounds, the model still repeatedly recalls documents that contribute little new evidence (as shown in Figure~\ref{fig:Retrieval-Intent Residual Computation}(a)). To enable the retriever to discover new evidences for next-hop reasoning, we propose a retrieval-intent residual computation method. This method strips away the semantic components of the query embedding that have been referred by the current clue, so that the residual can better guide the retriever to capture the clue required for the next hop (as shown in Figure~\ref{fig:Retrieval-Intent Residual Computation}(b)).

Let \(Q^{\mathrm{embed}}\) denote the current query embedding, let \(e\) denote the
textual bridging clue, and let \(e^{\mathrm{embed}}\) denote its embedding. We compute
the retrieval-intent residual by subtracting from \(Q^{\mathrm{embed}}\) its component
along \(e^{\mathrm{embed}}\) as:
\vspace{-0.1cm}
\begin{equation}
	Q_{\mathrm{res}}^{\mathrm{embed}}
	=
	Q^{\mathrm{embed}}
	-
	\frac{Q^{\mathrm{embed}}\cdot e^{\mathrm{embed}}}
	{\|e^{\mathrm{embed}}\|^2}e^{\mathrm{embed}}
	\label{eq-res}
\end{equation}

The resulting vector \(Q_{\mathrm{res}}^{\mathrm{embed}}\) captures the components of
\(Q^{\mathrm{embed}}\) that are orthogonal to the embedding of bridging-clue. This projection has a clear geometric interpretation and serves as a practical reranking signal rather than a strict discriminator of natural‑language semantics.  %

\subsection{Semantic Residual-based Reranking}
\label{subsection:reranking}
After retrieving candidate documents based on the bridging clue, the documents that are most relevant to the clue need to be identified, avoiding arbitrary drifting of evidence chains. To this end, a semantic residual-based reranking method is proposed to score documents using Formula \ref{eq:rerank}.
\vspace{-0.1cm}
\begin{equation}
\label{eq:rerank}
\begin{aligned}
S_{\text{rerank}}(c^{embed})
=
\lambda\,\operatorname{sim}(Q^{embed},c^{embed}) \\
+
(1-\lambda)\,\operatorname{sim}(Q_{\text{res}}^{\mathrm{embed}},c^{\mathrm{embed}})
\end{aligned}
\vspace{-0.1cm}
\end{equation}
where $\operatorname{sim}(\cdot,\cdot)$ computes similarity,
$Q^{\mathrm{embed}}$ is the original query embedding,
$Q_{\mathrm{res}}^{\mathrm{embed}}$ is the retrieval-intent residual,
$c^{\mathrm{embed}}$ is the candidate-document embedding, and
$\lambda\in[0,1]$ is the mixing coefficient between the two similarity terms.

This scoring method jointly captures the global semantic relevance to the original query and the targeted relevance to the retrieval-intent residual, thereby upweighting candidate documents that address the unresolved semantic components along the current retrieval path. %
Finally, the top-\(D\) most relevant documents are retained. Each of the retained documents is subsequently appended to the current evidence chain in parallel, yielding $D$ newly expanded reasoning paths.

\subsection{Scoring Mechanism for Evidence Chains}
\label{subsection:scoring}
In the stepwise reasoning stage, it is essential to estimate the reliability of expanded paths. We define a unified scoring function composed of a local node score (Definition \ref{def:localNodeScore}) and a cumulative path score (Definition \ref{def:cumulativeScore}). The scoring function jointly considers the global relevance to the original query and the guidance provided by the retrieval residual for the next hop.  %

\begin{definition}[Local Node Score]
\label{def:localNodeScore}
Given the original query embedding $Q^{\text{embed}}$, the retrieval-intent residual $RQ^{\text{embed}}$, a bridging clue $e$, and a candidate document $q$, the local node score of $q$ is defined as:
\vspace{-0.1cm}
\begin{equation}
	\label{eq:slocal}
	\begin{aligned}
		S_{\mathrm{local}}(q)
		=&\ \lambda_{\mathrm{vec}}\cos(Q^{\mathrm{embed}}, q^{\mathrm{embed}}) \\
		&+ \lambda_{\mathrm{intent}}\operatorname{ReLU}(R_Q^{\mathrm{embed}}\cdot q^{\mathrm{embed}}) \\
		&+ \lambda_{\mathrm{sym}}S_{\mathrm{sym}}(q,e).
	\end{aligned}
\end{equation}
 where \(\mathrm{Tok}(\cdot)\) denotes the token set obtained after
 lowercasing and punctuation stripping, and
 \(S_{\mathrm{sym}}(q,e)=
 \frac{|\mathrm{Tok}(e)\cap\mathrm{Tok}(q)|}
 {|\mathrm{Tok}(e)|}\)
 measures the explicit token coverage of the textual bridging clue \(e\)
 in document \(q\).
 $q^{embed}$ and $e^{embed}$ are the embeddings of $q$ and $e$. 
\end{definition}

The three terms keep the path aligned with the original query, favor documents matching the unresolved retrieval intent, and require explicit support for the current bridging clue.

\begin{definition}[Cumulative Score]
\label{def:cumulativeScore}
Let the current evidence chain is $D'$ with cumulative score $s'$. When a candidate document $q$ is appended to $D'$ to form a new path $D'' = D'.\operatorname{append}(q)$, the cumulative score of $D''$ is defined as:
\vspace{-0.1cm}
\begin{equation}
\operatorname{S}_{path}(D'')
=s'+ \gamma^{|D'|}\cdot S_{\text{local}}(q)
\label{eq:cscore}
\vspace{-0.1cm}
\end{equation}
where $\gamma \in (0,1)$ is the step discount factor, and $|D'|$ denotes the path depth.
\end{definition}

The cumulative scoring mechanism enables the search process to preferentially preserve reasoning trajectories that are continuously supported by high-quality local evidence. The discount term penalizes excessively long paths and mitigates the accumulation of uncertainty in deeper expansions. Ultimately, the cumulative scoring mechanism jointly considers the historical score of the path and the importance of the current document.

\section{Experiments and Analysis}
\label{section:experiments}
In this section, we first introduce the experimental setup (Section \ref{subsection:expSetup}), and then analyze the effectiveness of the internal-attention-guided mechanism (Section \ref{subsection:attentionAnalysis}). Subsequently, comparisons are conducted to evaluate the overall performance of our method (Section \ref{subsection:comparison}). %

\subsection{Experimental Setup}
\label{subsection:expSetup}
\textbf{Datasets:} Three representative multi-hop question answering datasets, 2Wiki \citep{ho2020multihopdataset}, MuSiQue \citep{trivedi2022musique}, and HotpotQA \citep{yang2018hotpotqa}, are selected as benchmark datasets.

\textbf{Baseline Methods:}
Three categories of retrieval method are selected as baselines. They are the sparse retrieval method BM25 \citep{robertson2009bm25}, the standard dense retrieval model GTR (T5-Large) \citep{ni2022gtr}, and strong large-scale embedding models, including Qwen3-Embedding-8B \citep{qwenteam2025qwen3embedding} and SFR-Embedding-Mistral-7B \citep{meng2024sfrembeddingmistral}. Our method is build on Qwen3-Embedding-4B.

\textbf{Implementation Details:}
By default, all experiments use Qwen3-Embedding-4B with the same probing configuration.
The maximum reasoning depth is fixed to $T=3$ for all three datasets. The weights in local node score
are $\lambda_{\mathrm{vec}}=8$, $\lambda_{\mathrm{intent}}=7$, and
$\lambda_{\mathrm{sym}}=20$. The reranking coefficient in
Eq.~\ref{eq:rerank} is $\lambda=8/(8+7)=0.533$, and the step discount is
$\gamma=0.90$. These settings are not retuned for individual datasets.
In Appendix~\ref{appendix:weight_sensitivity}, we analyze the hyperparameters in detail.

\textbf{Evaluation Metric:}
For each query and each $F\in\{5,10,15,20\}$, IN-Retriever retains the
top-$F$ reasoning paths and deduplicates their documents. Let $D_F$
denote the resulting document count. Each baseline then returns its
top-$D_F$ documents for the same query. Recall@$F$ therefore measures
supporting-document coverage under $F$ paths. Appendix~\ref{appendix:path_f1}
reports the complementary document-level F1 analysis under the same
alignment protocol.

\subsection{Attention Analysis}
\label{subsection:attentionAnalysis}

The effectiveness of IN-Retriever is grounded in a fundamental hypothesis: the elite attention heads of a pre-trained embedding model imply the clues required for reasoning. In this subsection, we analyze the model's layer-wise attention weights, verify the stability of the elite attention heads that play the dominant role in locating reasoning clues, and visualize two cases for understanding.

\subsubsection{Layer-wise Sensitivity}
\label{subsubsec:layerWiseSensitivity}
\begin{figure*}[htbp]
\vspace{-0.1cm}
  \centering
  \begin{subfigure}[]{0.3\textwidth}
    \includegraphics[width=\linewidth]{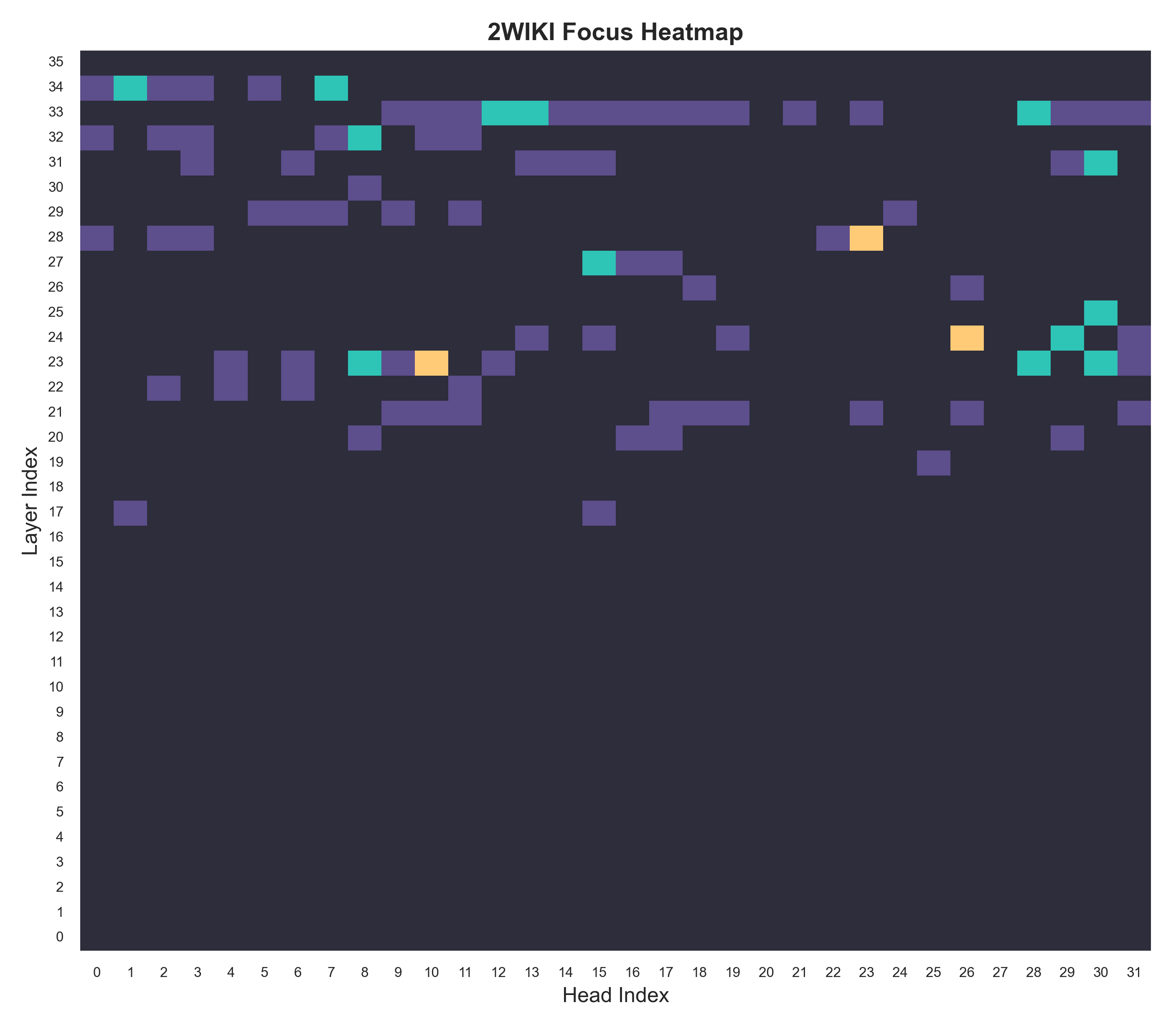}
    \caption{2Wiki}
    \label{fig:LayerBFS-2Wiki}
  \end{subfigure}
  \begin{subfigure}[]{0.3\textwidth}
    \includegraphics[width=\linewidth]{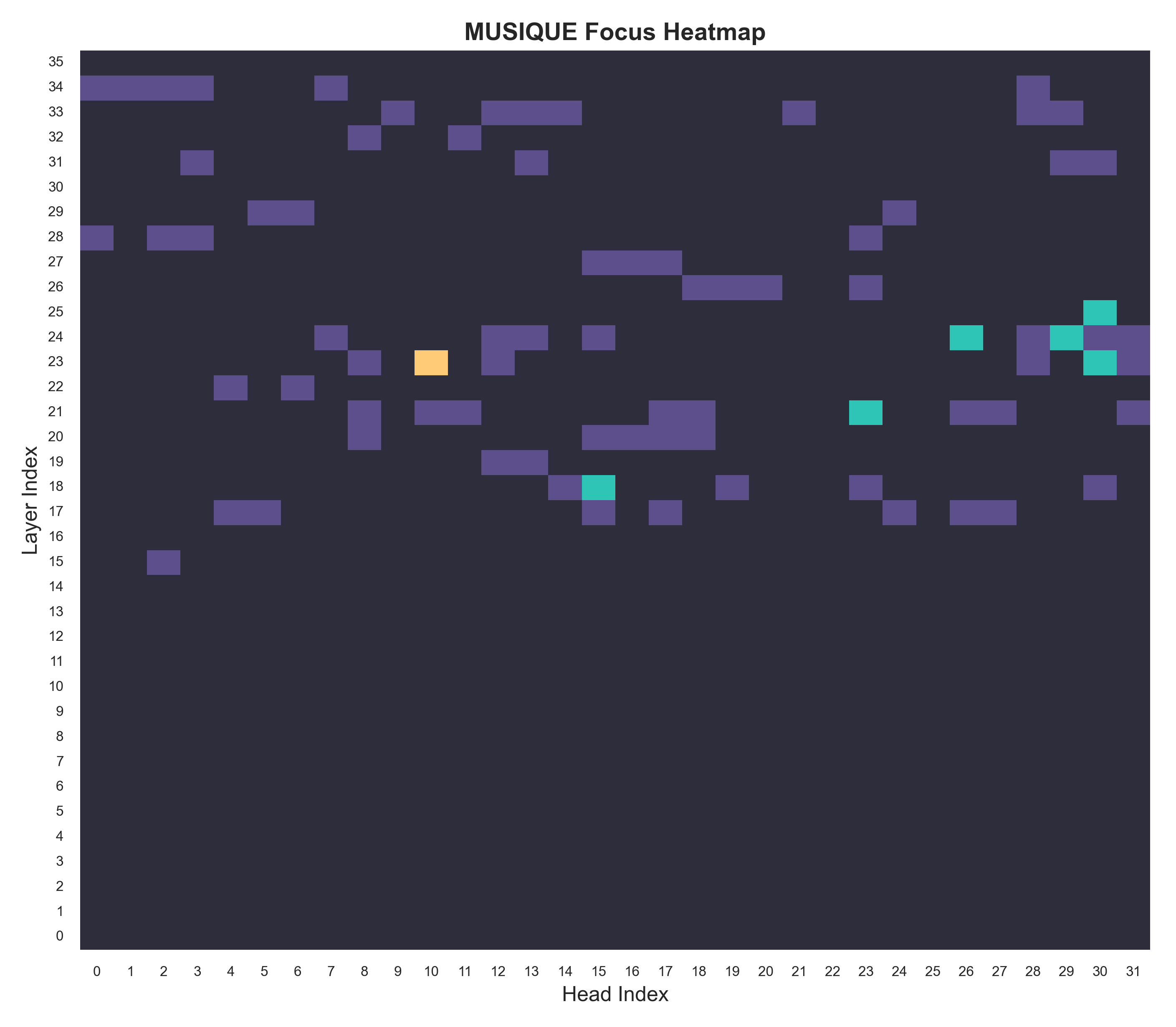}
    \caption{MuSiQue}
    \label{fig:LayerBFS-Musique}
  \end{subfigure}
  \begin{subfigure}[]{0.37\textwidth}
    \includegraphics[width=\linewidth]{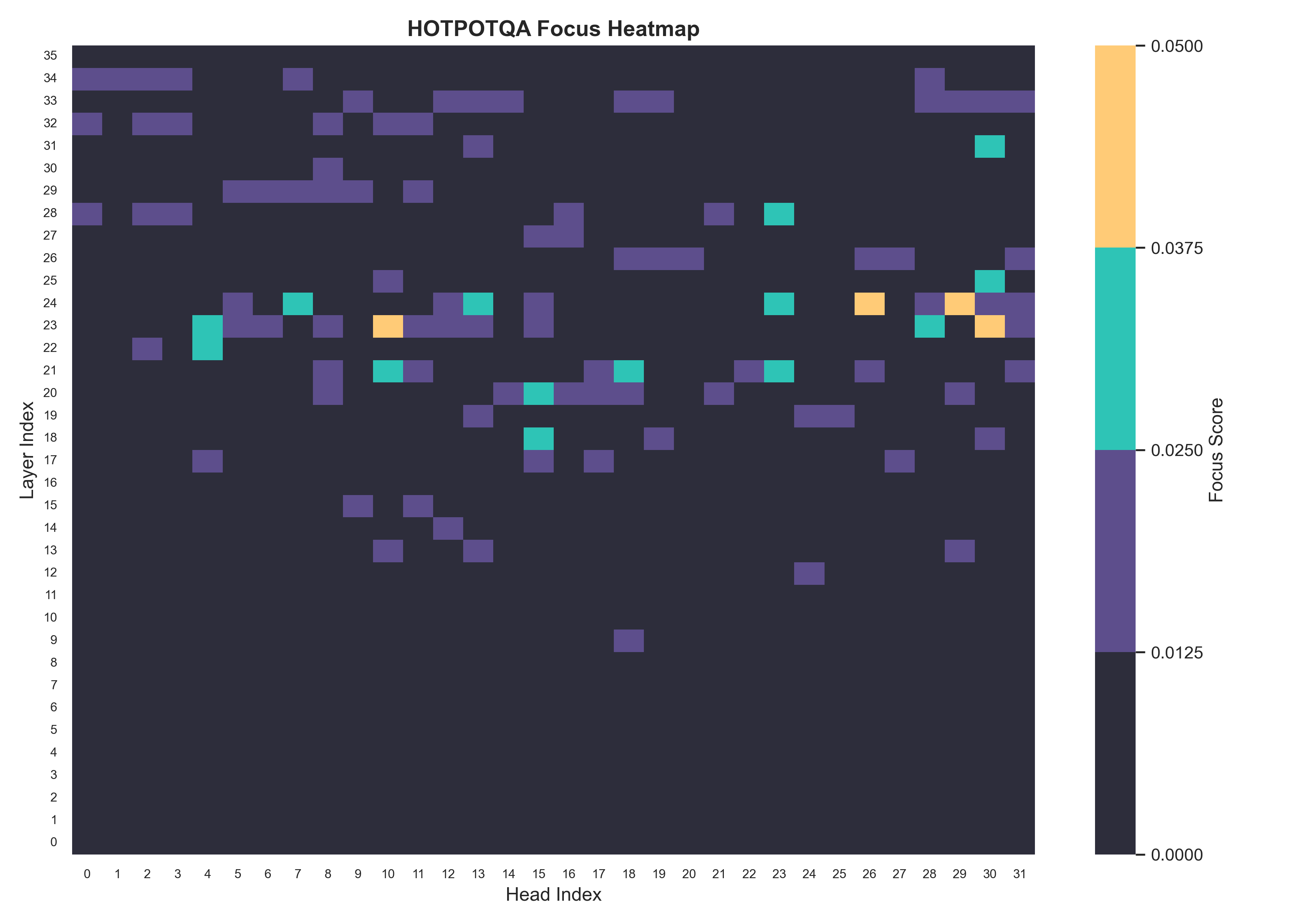}
    \caption{HotpotQA}
    \label{fig:LayerBFS-HotpotQA}
  \end{subfigure}
\vspace{-0.1cm}
  \caption{Heatmaps of Bridge Focus Score (BFS) on 2Wiki, MuSiQue, and HotpotQA.}
  \label{fig:LayerBFS}
\end{figure*}

To investigate the layer-wise pattern of the attention mechanism, we randomly sampled 400 query-context pairs from 2Wiki, fed them into the model, and output attention weights. The heat map of bridge focus score (BFS) across attention heads is visualized as Figure \ref{fig:LayerBFS-2Wiki}. The figure shows that attention heads with high BFS values are mainly concentrated in the middle-to-late layers of the model, revealing a sparse and clearly non-uniform spatial distribution. Similar results (as shown in Figure \ref{fig:LayerBFS-Musique} and Figure \ref{fig:LayerBFS-HotpotQA}) are also supported in the other datasets.

\begin{figure*}[t]
\vspace{-0.1cm}
	\centering
	\begin{subfigure}[t]{0.48\textwidth}
		\centering
		\includegraphics[width=\linewidth]{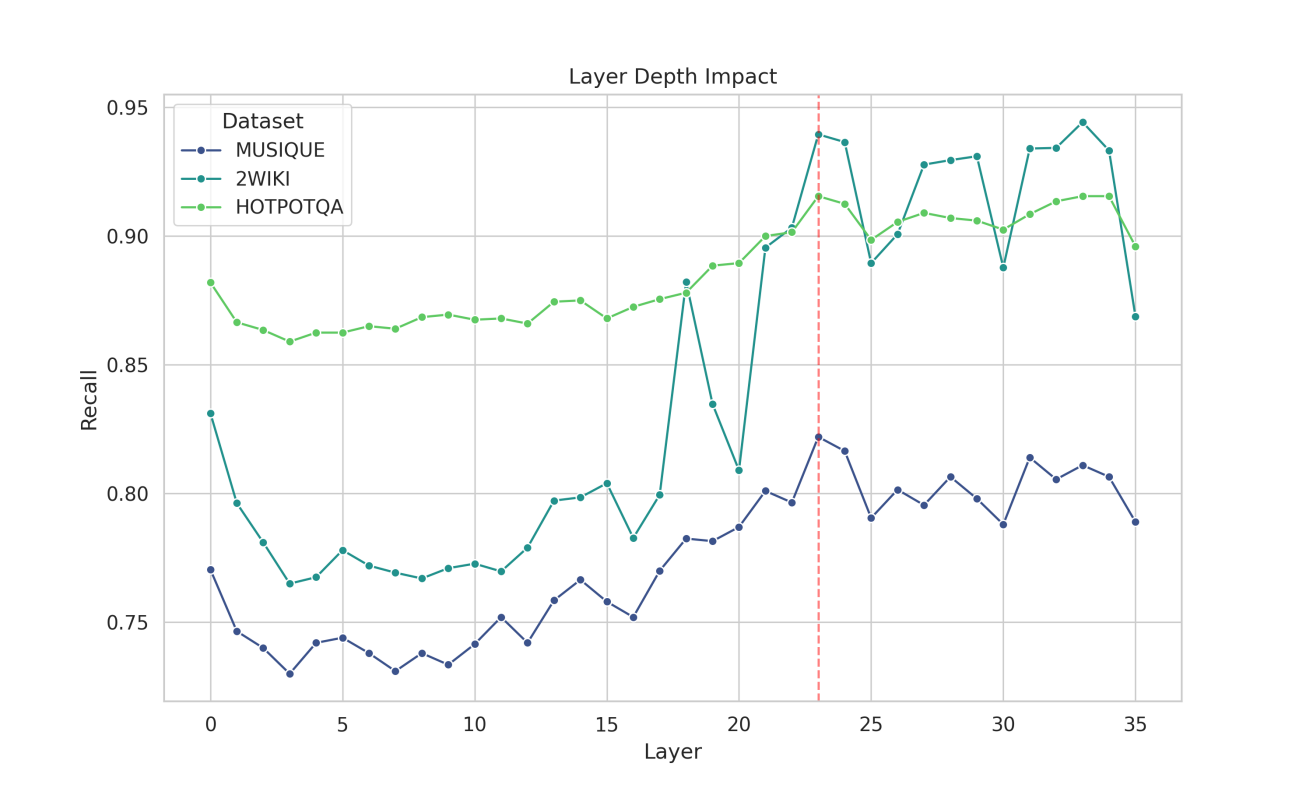}
		\caption{Recall@20 vs. attention layer depth.}
		\label{fig:layerwise_sensitivity_b}
	\end{subfigure}
	\hfill
	\begin{subfigure}[t]{0.48\textwidth}
		\centering
		\includegraphics[width=\linewidth]{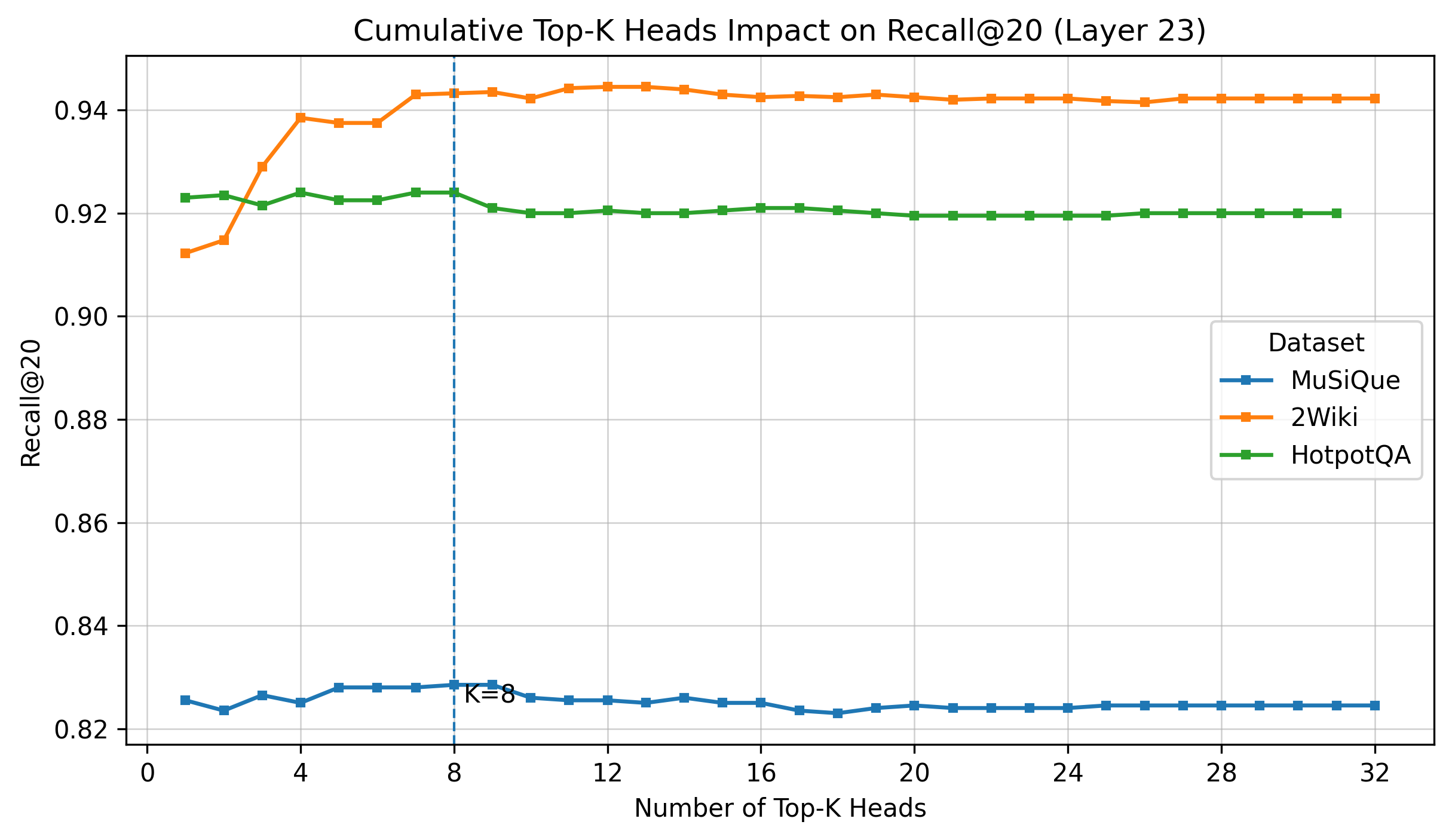}  %
		\caption{Cumulative impact of top-$H$ attention heads.}
		\label{fig:CumulativeImpact}
	\end{subfigure}
\vspace{-0.1cm}
	\caption{Layer-wise and head-wise analysis for selecting elite attention heads on 2Wiki, MuSiQue, and HotpotQA.}
	\label{fig:attention_selection}
\end{figure*}

Following the setting, we extracted the top-8 attention heads with the highest BFS values in each layer and evaluated the ability to retrieve the correct next-hop documents via the bridging clues. The results are shown in Figure \ref{fig:layerwise_sensitivity_b}.

Figure \ref{fig:layerwise_sensitivity_b} shows that the recall exhibit a consistent trend across all three datasets. The recall first decreases in layers 0 to 3, then remains relatively low and stable from layers 4 to 12. It subsequently rises gradually from layers 13 to 23 with fluctuations. Finally, the recall reaches its peak at layer 23, followed by fluctuations across layers 24–34. In the last layer (layer 35), the recall exhibits a distinct decline. This layer-wise trend is broadly consistent with earlier BERT probing studies that reported lower-layer surface or lexical information and higher-layer semantic information \citep{jawahar2019bertstructure,tenney2019bertpipeline}. \citet{niu2022bertpipeline} later found that a simple layer-wise pipeline lacks conclusive empirical support and that layer depth alone does not fully explain BERT's internal organization.

\subsubsection{Analysis of Elite Attention Heads}
Furthermore, taking layer 23 as the basis, we retrieve the next-hop document using the top-\(H\) heads with higher BFS scores. Figure \ref{fig:attention_selection}(b) shows the relationship between $H$ and the recall of next-hop document. Across all three datasets, the figure shows that the head exhibiting the highest BFS encapsulate the vast majority of the model's reasoning capacity. As \(H\) increases, the highest recall is achieved when $H=8$. On MuSiQue and 2Wiki, recall increases noticeably as the number of attention heads grows when $H<8$. When $H>8$, recall on 2Wiki largely plateaus, whereas recall on MuSiQue and HotpotQA shows a certain degree of decline. These results suggest that it is sufficient to focus on attention heads with the higher BFS values, while incorporating too many heads may instead introduce noise and degrade the performance. Taking both performance and stability into account, we finally select the attention heads specified by $\mathcal{H}_{\text{elite}}=\{10,30,8,28,31,6,4,9\}$ in layer $L=23$ as the elite attention heads for the Qwen3-Embedding-4B.

\subsubsection{Case Analysis}

\begin{figure*}[htbp]
\vspace{-0.2cm}
  \centering
  \begin{subfigure}[t]{0.48\textwidth}
    \centering
    \includegraphics[width=\linewidth]{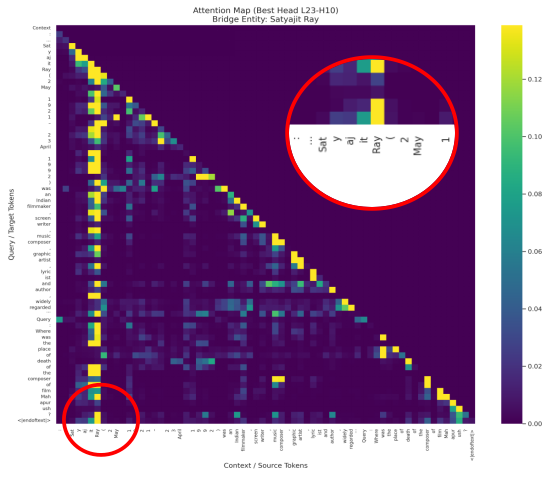}
    \caption{Case 1}
    \label{fig:AttentionCase1}
  \end{subfigure}
  \hfill
  \begin{subfigure}[t]{0.48\textwidth}
    \centering
    \includegraphics[width=\linewidth]{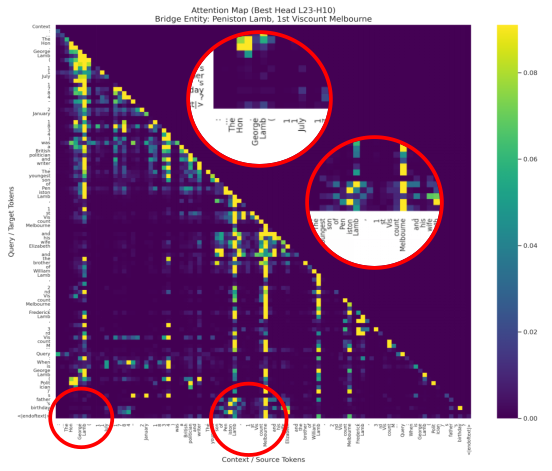}
    \caption{Case 2}
    \label{fig:AttentionCase2}
  \end{subfigure}
\vspace{-0.1cm}
  \caption{Case analysis of token-to-token in head 10, layer 23. (a) Case 1 shows persistent vertical attention bands. (b) Case 2 shows a dense semantic interference from homogeneous entities}
  \label{fig:AttentionCase}
\vspace{-0.1cm}
\end{figure*}

Focusing on head 10 at layer 23, we visualize token-level attention patterns of two representative cases. The weight matrix between the query and the context is shown in Figure \ref{fig:AttentionCase}. %

In Figure \ref{fig:AttentionCase1}, it exhibits distinct and persistent vertical bands. This geometric pattern indicates that the attention head consistently focuses on the same bridging clue across multiple query tokens, rather than distributing attentions diffusely over the broader context. In Figure \ref{fig:AttentionCase2}, it show a case with dense semantic interference, where the context contains multiple homogeneous named entities. Even in the presence of several same-type distractors, the attention remains strongly discriminative and aligns more concentratedly with the target clue. The visualization of the two cases further verifies that the selected attention head is not merely performing surface-level lexical matching, but instead provides a stable and interpretable signal for clue localization.

\subsection{Comparative Analysis}
\label{subsection:comparison}
The comparison results between IN-Retriever and retrieval baselines on the three
multi-hop datasets are shown in Table~\ref{tab:retrieval_comparison}. The quantitative
evaluation focuses on the retriever itself: whether it recalls supporting evidence under
the retrieval defined above.  %

\begin{table*}[t!]
\caption{Comparison of retrieval methods on MuSiQue, 2Wiki, and HotpotQA. $F$ denotes the number of retained IN-Retriever paths. Baseline methods select the corresponding $D_F$ deduplicated documents for each query comparisons. The best performances are \textbf{bolded}, and the second-best performances are \underline{underlined}.}
\label{tab:retrieval_comparison}
\vspace{-0.2cm}
\centering
\setlength{\tabcolsep}{4pt}
\resizebox{\textwidth}{!}{%
\begin{tabular}{lcccccccccccc}
\toprule
\multirow{2}{*}{\textbf{Retrieval Methods}} 
& \multicolumn{4}{c}{\textbf{MuSiQue}} 
& \multicolumn{4}{c}{\textbf{2Wiki}} 
& \multicolumn{4}{c}{\textbf{HotpotQA}} \\
\cmidrule(lr){2-5} \cmidrule(lr){6-9} \cmidrule(lr){10-13}
& \textbf{R@5} & \textbf{R@10} & \textbf{R@15} & \textbf{R@20}
& \textbf{R@5} & \textbf{R@10} & \textbf{R@15} & \textbf{R@20}
& \textbf{R@5} & \textbf{R@10} & \textbf{R@15} & \textbf{R@20} \\
\midrule
\multicolumn{13}{l}{\textit{Simple Baselines}} \\
BM25 
& 48.80 & 53.30 & 57.40 & 59.30 
& 64.50 & 68.43 & 70.17 & 70.97 
& 76.35 & 87.70 & 91.15 & 92.60 \\
GTR (T5-Large) 
& 53.50 & 59.65 & 64.15 & 67.05 
& 65.87 & 69.05 & 70.63 & 71.75 
& 78.50 & 84.15 & 86.70 & 88.10 \\
\midrule
\multicolumn{13}{l}{\textit{Large Embedding Models}} \\
SFR-Embedding-Mistral-7B 
& \underline{65.65} & \underline{72.55} & \underline{77.90} & \underline{79.75} 
& 73.95 & \underline{78.17} & \underline{79.73} & \underline{80.77} 
& \textbf{90.60} & \textbf{95.10} & \textbf{96.30} & \textbf{96.90} \\
Qwen3-Embedding-8B 
& 65.40 & 72.45 & 77.85 & 79.30 
& \underline{74.02} & 77.32 & 79.43 & 80.55 
& \underline{88.75} & \underline{93.30} & \underline{95.10} & 95.85 \\
\midrule
\multicolumn{13}{l}{\textit{Baseline RAG}} \\
Qwen3-Embedding-4B (Direct) 
& 63.80 & 70.20 & 75.35 & 77.95 
& 72.82 & 76.28 & 77.85 & 78.90 
& 86.10 & 92.15 & 93.90 & 94.80 \\
\midrule
\textbf{IN-Retriever} 
& \textbf{76.45} & \textbf{82.80} & \textbf{85.40} & \textbf{87.10}
& \textbf{89.10} & \textbf{94.33} & \textbf{95.27} & \textbf{95.57}
& 86.35 & 92.10 & 94.75 & \underline{96.70} \\
\bottomrule
\end{tabular}%
}
\vspace{-0.2cm}
\end{table*} 

As shown in Table \ref{tab:retrieval_comparison}, IN-Retriever with Qwen3-Embedding-4B as the backbone achieves the best results on R@5, R@10, R@15, and R@20 for the more challenging MuSiQue and 2Wiki datasets. Meanwhile, on HotpotQA, it outperforms direct retrieval baselines of similar scale, including BM25, GTR, and Qwen3-Embedding-4B, while remaining competitive with larger embedding models such as SFR-Embedding-Mistral-7B and Qwen3-Embedding-8B.

Taking R@20 as an example, comparing with simple baselines, BM25 and GTR-Large, our method improves R@20 by +27.80 and +20.05 on MuSiQue, by +24.60 and +23.82 on 2Wiki, and by +4.10 and +8.60 on HotpotQA. These results reveal the following two key observations: (1) explicitly leveraging internal attention signals to guide retrieval is effective for multi-hop evidence search; and (2) for multi-hop queries, relying solely on embedding-based document retrieval is insufficient. 

We further examine the recall of two larger embedding models, SFR-Embedding-Mistral-7B and Qwen3-Embedding-8B. First, despite their larger parameter scales, their performances do not apparently exceed that of Qwen3-Embedding-4B (Direct). This suggests that although larger embedding models are generally more favorable for document retrieval and can improve recall, such gains remain limited in multi-hop settings. Moreover, as shown in Table \ref{tab:retrieval_comparison}, except on HotpotQA, our method achieves the best recall on the other two datasets, which demonstrates that even when using Qwen3-Embedding-4B as the backbone, IN-Retriever can in most cases match or even surpass larger embedding model baselines. This further confirms the necessity of targeted multi-hop clue discovery and stepwise retrieval guided by such clues.

Finally, a comparison across datasets shows that the recall improvements of our method are more pronounced on MuSiQue and 2Wiki, while the gains on HotpotQA are relatively limited. This is consistent with the analysis in \citet{min2019compositional}, which suggests that a substantial fraction of HotpotQA queries can be resolved through reasoning shortcuts, with answers obtainable from a single document. In contrast, 2Wiki and MuSiQue rely more on bridging evidence across documents. In such tasks, the system must not only retrieve documents that are superficially relevant to the original question, but also progressively identify the bridging clues that connect different pieces of evidence. Compared with methods that rely on a single query vector alone, IN-Retriever demonstrates clearer advantages in multi-hop retrieval and is better suited to retrieval tasks that require progressively locating multi-hop evidence.

\section{Conclusion}
\label{section:conclusion}
Establishing a coherent and complete evidence chain remains a core challenge for multi-hop QA tasks. In this study, we propose a training-free Intrinsic Navigating-based non-generative Retrieval method (IN-Retriever). The method directly extracts bridging clues from the internal attention of the embedding model and uses them to retrieve candidate documents. %

Based on Qwen3-Embedding-4B, we first analyze the BFSs across attention heads, revealing clear layer-wise sensitivity and stable distribution patterns among high-BFS heads. Then, the experiments verify the consistency of the elite attention heads across datasets. Comparative experiments indicate that IN-Retriever is effective at locating bridging clues and retrieving supporting evidence, and that attention weights can be directly leveraged to guide multi-hop retrieval. 

Overall, our findings suggest that pre-trained embedding models contain internal attention signals that support multi-hop retrieval in a non-generative manner. This study offers a practical idea for bridging internal model signals with external retrieval in complex reasoning tasks.

\section*{Limitations}
Although IN-Retriever shows strong gains on MuSiQue and 2Wiki and remains competitive on HotpotQA, the conclusions of this work should be interpreted within a limited scope. Our results show that intrinsic attention in embedding models can provide a useful signal for bridge-clue discovery in multi-hop retrieval, but they do not provide a causal explanation of how embedding models perform multi-hop reasoning. Attention is used here as a retrieval signal rather than as an internal reasoning process.

The orthogonal projection used in retrieval‑intent residual computation has a clear geometric interpretation and is empirically supported by ablation studies, yet noise in bridging‑clue representations and the diversity of entity expressions may reduce the reliability of residual guidance.

IN-Retriever is training-free. However, it requires white-box access to internal attention weights and a one-time calibration step to identify elite attention heads for a given backbone. Although we observe similar sparse BFS patterns in other architectures, comprehensive retrieval evaluations across diverse embedding backbones, closed-source retrieval APIs, multilingual scenarios and domain-specific models have not yet been conducted.

The current clue extraction procedure also relies on several lightweight hand-designed steps, including sentence filtering, peak-token selection, boundary expansion, lexical support scoring, and fixed search budgets. These choices make the method simple and interpretable, but they may be less robust when bridging clues are weakly lexicalized, highly paraphrastic, distributed across multiple sentences, or expressed through temporal, numerical, comparative, or implicit relational constraints. The method is therefore most naturally suited to entity- or noun-phrase-centric multi-hop questions, and its behavior on longer or more highly branching evidence chains remains to be studied.

Finally, our evaluation focuses on retrieval quality rather than end-to-end generation. This matches the main contribution of the paper, but higher Recall@F does not necessarily imply proportional improvements for every downstream generator. In addition, IN-Retriever returns a dynamic number of deduplicated documents and requires iterative attention probing, multi-step retrieval, and reranking, which introduce more inference overhead than single-shot dense retrieval. Future work should study fixed-budget evaluation, end-to-end RAG performance, and latency-quality trade-offs in realistic deployments.

\bibliography{IN-Retriever}

\clearpage
\appendix
\section*{Appendix}
\label{appendix}

\section{Related Work}
In this section, we review researches closely related to our study. They are document retrieval methods for multi-hop queries and some studies about attention heads in embedding method.

\paragraph{Document Retrieval for Multi-Hop Queries.}
RAG helps alleviate issues such as hallucination and outdated information in LLMs. However, in practical scenarios, complex multi-hop queries often fail to require sufficient evidence documents for generation. Existing document retrieval methods for multi-hop queries within the RAG framework are broadly classified into three paradigms: (1) using dense retrieval to improve the recall of multi-hop evidence; (2) explicitly generating intermediate reasoning steps to plan subsequent retrieval paths; and (3) leveraging external explicit structures to guide reasoning.

To improve the recall, early research mainly focused on dense retrieval-based methods. Classical architectures, such as DPR \citep{karpukhin2020dpr} and ANCE \citep{xiong2021annncl}, employ dual encoders to map queries and documents into a shared latent space, attempting to capture semantic relevance through similarity matching. GTR shows that scaling a single-vector dual encoder can substantially improve retrieval performance \citep{ni2022gtr}. ColBERTv2 retains token-level representations and applies late interaction for finer-grained query--document matching \citep{santhanam2022colbertv2}.

To bridge the logical gap, recent works tend to explicitly generate intermediate reasoning steps and then plan subsequent retrieval paths. For example, IRCoT \citep{trivedi2023interleaving} interleaves chain-of-thought (CoT) reasoning with retrieval steps and generates reasoning plans using LLM. ReAct \citep{yao2023react} introduces an "action-observation" loop to dynamically adjust search strategies. Self-RAG \citep{asai2024selfrag} further incorporates a self-reflection mechanism to assess retrieval quality. LLMs may over-commit to early mistakes during sequential generation, allowing hallucinations to snowball \citep{zhang2023sirenssong}.

The third category of retrieval methods leverages \emph{external explicit structures} to provide reasoning constraints. For example, GraphRAG \citep{edge2024graphrag} builds an LLM-derived entity graph and hierarchical community summaries during indexing, then generates partial and final responses from those summaries for corpus-level questions. Following this direction, recent studies have explored more sophisticated graph structures to improve multi-hop reasoning. \citet{yang2026hierarchicalkgqa} combine hierarchical knowledge graphs with RAG through graph convolutions that aggregate local and global relations for complex queries. \citet{chen2026topicgraph} propose a Topic-Graph-based method for long-document, enhancing cross-paragraph evidence association by activating relevant sentences and their neighborhoods along the graph structure. \citet{chen2026dairygoatqa} use relational paths as the basic unit and exploit explicit path constraints to guide multi-hop evidence selection in domain-specific QA. Overall, these methods provide clear logical and structural constraints for multi-hop retrieval, but their practical use is often limited by high construction and maintenance costs. They are not only highly sensitive to the quality of upstream text preprocessing, but also incur substantial update costs when external knowledge evolves dynamically.

The comparisons of the above methods are summarized in Table~\ref{tab:qualitative_comparison}.  
The comparisons provides a qualitative comparison of paradigm-level design properties. We do not compare them in our experiments, since their final performance is influenced by multiple factors including generation, prompting, and graph construction.
\begin{table*}[htbp]
\centering
\footnotesize
\setlength{\tabcolsep}{4pt}
\caption{Comparison of representative multi-hop retrieval methods.}
\label{tab:qualitative_comparison}
\begin{tabular}{
L{2.8cm}
L{2.8cm}
C{1.5cm}
C{2.0cm}
C{2.5cm}
C{2.2cm}
}
\toprule
\textbf{Paradigm} &
\textbf{Representative Methods} &
\textbf{\makecell[c]{Non-\\Generative}} &
\textbf{\makecell[c]{No External\\Graph/Path}} &
\textbf{\makecell[c]{Stepwise Retrieval\\Guidance}} &
\textbf{\makecell[c]{Robust to Error\\Propagation}} \\
\midrule
Dense Retrieval &
\makecell[l]{DPR, ANCE, \\GTR-Large} &
\cmark & \cmark & \xmark & \cmark \\
\midrule
Generative Reasoning-Guided Retrieval &
\makecell[l]{IRCoT, ReAct,\\Self-RAG} &
\xmark & \cmark & \cmark & \xmark \\
\midrule
External-Structure-Guided Retrieval &
\makecell[l]{GraphRAG,\\Hierarchical\\KG-RAG,\\Topic-Graph RAG} &
\xmark & \xmark & \cmark & \cmark \\
\midrule
\rowcolor{gray!10}
\textbf{\makecell[l]{}} &
\textbf{IN-Retriever} &
\textbf{\cmark} & \textbf{\cmark} & \textbf{\cmark} & \textbf{\cmark} \\
\bottomrule
\end{tabular}
\end{table*}

\paragraph{Attention Heads of Embedding Models.}
Recently, studies in mechanistic interpretability try to open the black box of Transformer models. For example, study on induction heads \citet{olsson2022inductionheads} reveals that some attention heads can automatically locate previously observed patterns in the context and copy the information that follows, thereby supporting sequence-level pattern continuation and information transfer. By contrast, ROME \citep{meng2022editing} uses causal interventions to show that MLP layers are key loci for storing factual knowledge. These findings motivate our investigation of internal model signals for guiding external document retrieval during inference.

\section{Supplementary Experiments}
\label{appendix:experiments}

\subsection{Reproducibility Details}
\label{appendix:reproducibility}
We use ChromaDB as the vector datastore. Collections are named
\texttt{\{dataset\}\_\{split\}\_\{sample\_size\}} and use an HNSW index with L2
distance. For 2Wiki and HotpotQA, sentences sharing the same title are concatenated
into one document; for MuSiQue, paragraphs sharing the same title are aggregated.
All documents are encoded with Qwen3-Embedding-4B before insertion into the collection.

The candidate samples for $\mathcal{D}_{\mathrm{calib}}$ are drawn from the 2Wiki
training split. We construct 2,000 candidate samples of the compositional type, each containing at least two complete evidence
triples. The object entity in the first-hop evidence triple is used as the bridging
clue $E$. During filtering, $E$ and context $C$ are lowercased, and a sample is retained
only when $E$ appears as a contiguous string in $C$. BFS computation first uses exact
token-sequence matching to locate $E$ and falls back to its lowercased token sequence
when needed. Only samples producing valid BFS scans are aggregated to compute each
attention head's final BFS.
\subsection{Stability of Elite Attention Heads}

To verify the stability of elite attention heads, we design an experiment to examine whether elite attention head remains consistent across different datasets. To this end, we construct evaluation subsets from 2Wiki and MuSiQue. For 2Wiki, we retain only samples labeled as compositional and containing complete evidence triples. For MuSiQue, we use question decomposition annotations to extract samples with explicit bridging dependencies. Finally, we select 2,000 high-confidence samples from each dataset.  %

In the experiment, we extract the top-8 attention heads according to BFS score in layer $L=23$, denoted as $\mathcal{H}_{\text{2Wiki}}$ and $\mathcal{H}_{\text{MuSiQue}}$, respectively. The Jaccard similarity coefficient is used as the evaluation metric, as defined in Formula \ref{eq:jaccard}.
\begin{equation}
J(\mathcal{H}_{\text{ds}}, \mathcal{H}_{\text{elite}})
=
\frac{|\mathcal{H}_{\text{ds}} \cap \mathcal{H}_{\text{elite}}|}
{|\mathcal{H}_{\text{ds}} \cup \mathcal{H}_{\text{elite}}|}
\label{eq:jaccard}
\end{equation}
where \(ds \in \{\text{2Wiki}, \text{MuSiQue}\}\).

\begin{figure}[htbp]
  \centering
  \includegraphics[width=\columnwidth]{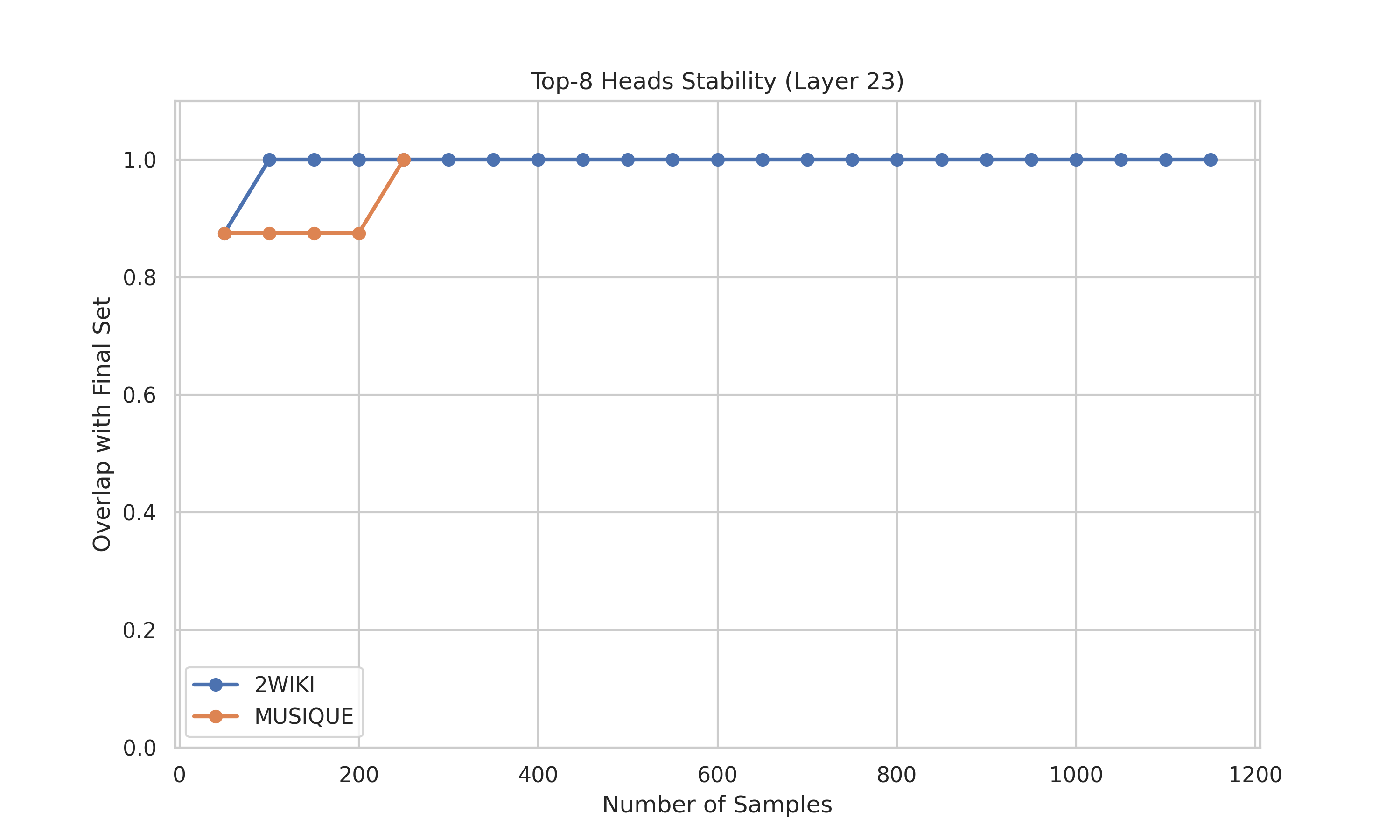}
  \caption{Overlap ratio with the elite attention heads as the number of calibration samples increases on 2Wiki and MuSiQue. }
  \label{fig:eliteOverlap}
\end{figure}

The experimental results are shown in Figure \ref{fig:eliteOverlap}. The figure shows that, as the number of samples increases, the top-\(H\) attention heads identified from partial data subsets eventually align perfectly with the elite attention heads $\mathcal{H}_{\text{elite}}$. Specifically, on 2Wiki, complete overlap is achieved when the sample size reaches 100. Even on MuSiQue, full overlap is achieved once the sample size reaches 250. The results indicate that elite attention heads are consistent across different datasets, and exhibit a certain degree of cross-task stability instead of overfitting to a specific dataset. Second, this overlap converges rapidly even with only a small number of calibration samples, indicating that the calibration cost is relatively low.

\subsection{Beam Width Analysis}
\label{sec:appendix}
IN-Retriever adopts the beam search mechanism, which allows it to maintain multiple potential reasoning trajectories simultaneously, thereby enabling the model to tolerate local uncertainty. In practice, the number of candidate paths (beam width, denoted by $B$) may affect the final retrieval results. In this subsection, we evaluate both retrieval effectiveness and the number of deduplicated documents in the final output under different values of $B$, in order to guide the selection of $B$.

\begin{figure}[htbp]
  \centering
  \includegraphics[width=\columnwidth]{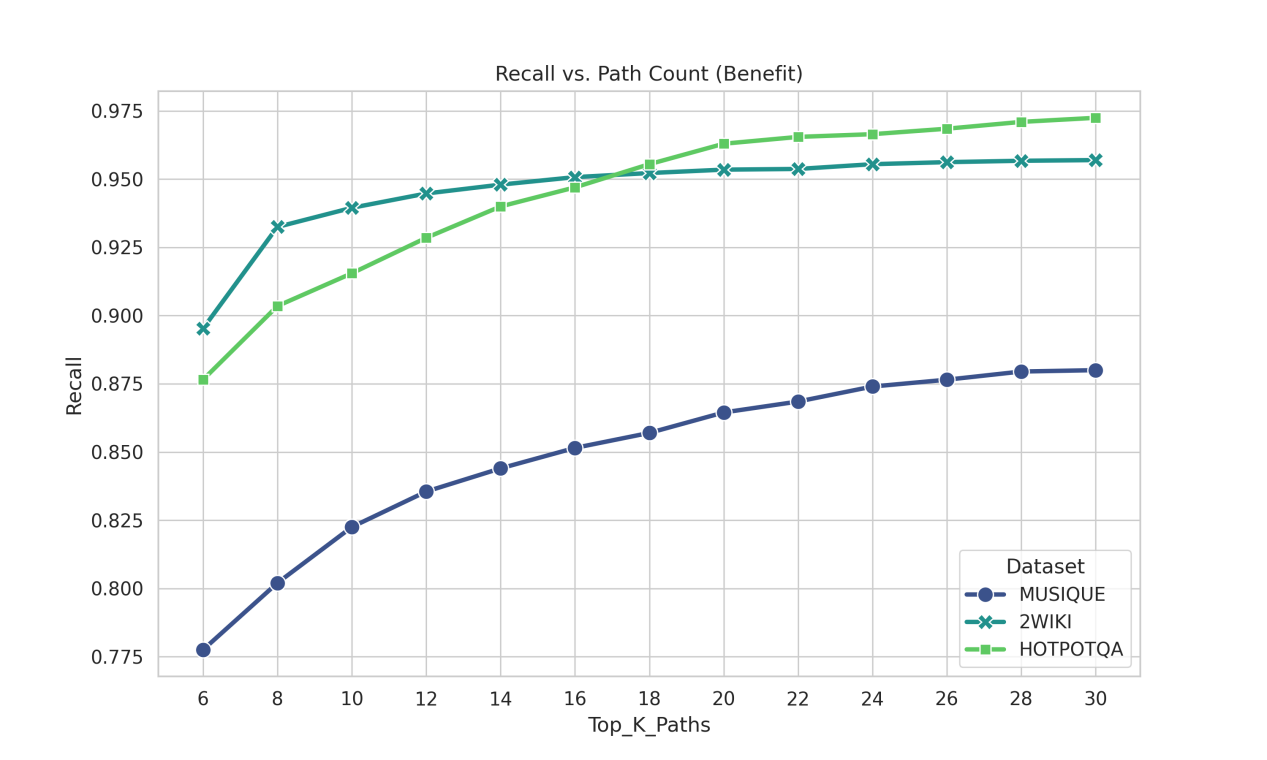}
  \caption{Curves of Recall@20 vs. beam widths $B$ on MuSiQue, 2Wiki, and HotpotQA}
  \label{fig:appendix_recall_vs_m}
\end{figure}

Figure \ref{fig:appendix_recall_vs_m} shows the curves between \(B\) and Recall@20 on three datasets. The curves exhibit an evident elbow between $B=8$ and $B=10$, beyond which the marginal gains become very limited. For example, Recall@20 on 2Wiki reaches 0.93 at $B=8$, and subsequent improvement slowed down apparently. The results suggest that tracking the top-\(B\) highest-confidence paths is sufficient to cover the vast majority of critical evidence. Accordingly, we set $B=8$ in the experiments.

\begin{figure}[htbp]
  \centering
  \includegraphics[width=\columnwidth]{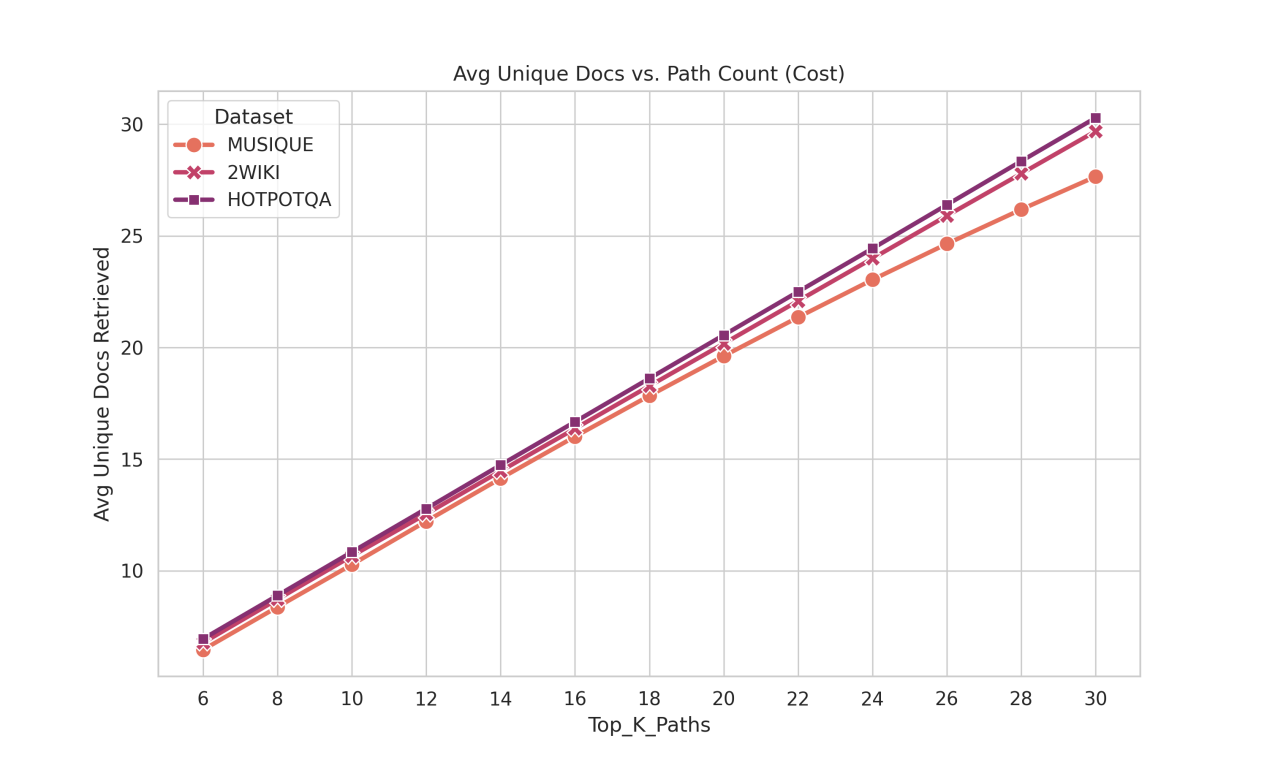}
  \caption{Average number of deduplicated documents versus path number $B$ on 2Wiki, MuSiQue, and HotpotQA.}
  \label{fig:appendix_docs_vs_k}
\end{figure}

Figure \ref{fig:appendix_docs_vs_k} shows the relation between the number of selected paths $B$ and the number of deduplicated documents. As shown in the figure, although each path may contain multiple documents, the aggregate document count does not scale multiplicatively with $B$; instead, it exhibits an approximately 1:1 growth trend with $B$. The result indicates that even though IN-Retriever retains more paths during reasoning, the number of documents does not diverge arbitrarily, but concentrates within a limited range.

The relation between path number and the size of documents further supports the selection of $B=8$. Real effective reasoning paths tend to converge within the top-8 paths and point to the same correct answer. Additional paths beyond this range mostly diverge into noisier regions. Therefore, limiting the number of paths in IN-Retriever is not only a matter of computational efficiency, but also an effective way to filter out long-tail noise and reduce reasoning drift.

\subsection{Path-Level F1 under Path Budget \(B\)}
\label{appendix:path_f1}

In contrast to fixed-document-budget retrieval, IN-Retriever returns a variable number of deduplicated documents for each query. A flexible evaluation metric is helpful for evaluating retrieval quality under path-aware evidence aggregation. Therefore, we use $F1$ at $B$ paths as an additional metric to further quantify the retrieval quality of IN-Retriever. 

For each query, the final retrieval documents are collected from $B$ paths. Let the corresponding gold document set be $Relevant$, and the set of documents finally returned by IN-Retriever be $Retrieved$ at $B$ paths. The number of correctly matched documents $Hit@B$, $Precision@B$, $Recall@B$, and $F1@B$ at $B$ paths are defined as Formula \ref{eq:hit}, \ref{eq:precision}, \ref{eq:recall}, and \ref{eq:F1}, where $B$ refers to $B$ paths.
\begin{equation}
\label{eq:hit}
Hit@B = |Relevant \cap Retrieved@B|.
\end{equation}
\begin{equation}
\label{eq:precision}
Precision@B
=
\frac{Hit@B}{|Retrieved|}
\end{equation}
\begin{equation}
\label{eq:recall}
Recall@B
=
\frac{Hit@B}{|Relevant|}
\end{equation}
\begin{equation}
\label{eq:F1}
F1@B
=
2 \times
\frac{Precision@B \times Recall@B}
{Precision@B + Recall@B}
\end{equation}

Assuming that the top-\(B\) paths contain \(R_B\) deduplicated documents, we extract the same number \(R_B\) of documents from each baseline and compute document-level \(F1\) under this aligned document budget. This alignment ensures that all compared methods are evaluated under approximately comparable reading length and downstream computational budget, thereby reducing the possibility that performance differences arise merely from returning more documents. The experimental results are shown in Table~\ref{tab:retrieval_comparison_f1}.

\begin{table*}[!ht]
	\caption{Comparison of retrieval methods in terms of document-level \(F1\) under different path budgets \(B\). The columns \(B=5,10,15,20\) denote the number of selected reasoning paths used to construct the retrieved document set. Each reported value is the average \(F1\) over all queries. The best performances are \textbf{bolded}, the second-best performances are \underline{underlined}, and IN-Retriever is highlighted in gray.}
	\label{tab:retrieval_comparison_f1}
	\centering
	\setlength{\tabcolsep}{4pt}
	\renewcommand{\arraystretch}{1.1}
	
	\resizebox{\textwidth}{!}{%
		\begin{tabular}{lcccccccccccc}
			\toprule
			\multirow{2}{*}{\textbf{Retrieval Methods}} 
			& \multicolumn{4}{c}{\textbf{MuSiQue}} 
			& \multicolumn{4}{c}{\textbf{2Wiki}} 
			& \multicolumn{4}{c}{\textbf{HotpotQA}} \\
			\cmidrule(lr){2-5} \cmidrule(lr){6-9} \cmidrule(lr){10-13}
			& \textbf{\(B=5\)} & \textbf{\(B=10\)} & \textbf{\(B=15\)} & \textbf{\(B=20\)}
			& \textbf{\(B=5\)} & \textbf{\(B=10\)} & \textbf{\(B=15\)} & \textbf{\(B=20\)}
			& \textbf{\(B=5\)} & \textbf{\(B=10\)} & \textbf{\(B=15\)} & \textbf{\(B=20\)} \\
			\midrule

\multicolumn{13}{l}{\textit{Simple Baselines}} \\
BM25
& 25.14 & 16.71 & 12.98 & 10.68
& 36.34 & 24.31 & 18.36 & 14.75
& 38.17 & 27.43 & 20.63 & 16.43 \\
GTR (T5-Large)
& 27.70 & 18.84 & 14.58 & 12.12
& 37.01 & 24.43 & 18.33 & 14.76
& 39.30 & 26.33 & 19.62 & 15.64 \\
\midrule

\multicolumn{13}{l}{\textit{Large Embedding Models}} \\
SFR-Embedding-Mistral-7B
& \underline{33.94} & \underline{22.92} & \underline{17.73} & \underline{14.45}
& \underline{41.82} & \underline{27.79} & \underline{20.82} & \underline{16.72}
& \textbf{45.55} & \textbf{29.82} & \textbf{21.83} & \textbf{17.21} \\
Qwen3-Embedding-8B
& 33.81 & 22.88 & \underline{17.73} & 14.37
& 41.81 & 27.44 & 20.69 & 16.62
& \underline{44.55} & \underline{29.25} & 21.55 & 17.02 \\
\midrule

\multicolumn{13}{l}{\textit{Baseline RAG}} \\
Qwen3-Embedding-4B (Direct)
& 33.04 & 22.16 & 17.15 & 14.13
& 41.08 & 27.01 & 20.23 & 16.24
& 43.21 & 28.87 & 21.28 & 16.83 \\
\midrule

\rowcolor[HTML]{EFEFEF}
\textbf{IN-Retriever}
& \textbf{40.16} & \textbf{26.33} & \textbf{19.60} & \textbf{15.90}
& \textbf{50.81} & \textbf{34.49} & \textbf{25.64} & \textbf{20.38}
& 43.84 & 29.03 & \underline{21.56} & \underline{17.20} \\
\bottomrule
\end{tabular}%
}
\end{table*}

As shown in Table~\ref{tab:retrieval_comparison_f1}, IN-Retriever achieves the best F1 scores on MuSiQue and 2Wiki across all evaluated values of $B$, indicating that the proposed method improves not only evidence recall but also the balance between retrieved evidence coverage and retrieval compactness. On HotpotQA, IN-Retriever remains competitive but does not outperform SFR-Embedding-Mistral-7B, which is consistent with the relatively weaker multi-hop dependency of this dataset. These results further support that IN-Retriever is more advantageous on datasets where evidence must be connected through explicit bridging clues.

\subsection{Normalized Weight Sensitivity Analysis}
\label{appendix:weight_sensitivity}

With the total local-score weight fixed at 35, we normalize the three
weights as
$(\alpha_v,\alpha_i,\alpha_s)
=(\lambda_{\mathrm{vec}},\lambda_{\mathrm{intent}},
\lambda_{\mathrm{sym}})/35$.
We vary $\alpha_v$ and $\alpha_i$ over
$\{0.0,0.2,0.4,0.6,0.8,1.0\}$, set
$\alpha_s=1-\alpha_v-\alpha_i$, and retain only combinations for which
all three normalized weights are non-negative.
Table~\ref{tab:weight_sensitivity_2wiki} reports Path Recall@20 on the
2Wiki validation set.

\begin{table}[!htb]
	\centering
	\caption{Fixed-total weight sensitivity on 2Wiki (Path Recall@20);
		dashes denote infeasible settings.}
	\label{tab:weight_sensitivity_2wiki}
	\small
	\setlength{\tabcolsep}{2pt}
	\renewcommand{\arraystretch}{1.08}
	\begin{tabular*}{\columnwidth}
		{@{\extracolsep{\fill}}c*{6}{r}@{}}
		\toprule
		$\alpha_v\backslash\alpha_i$
		& 0.0 & 0.2 & 0.4 & 0.6 & 0.8 & 1.0 \\
		\midrule
		0.0
		& 85.28 & 94.83 & 94.17 & 93.95 & 93.70 & 0.00 \\
		0.2
		& 95.73 & 95.80 & 95.90 & 95.58 & 90.20 & -- \\
		0.4
		& 95.80 & 95.93 & 95.93 & 91.27 & -- & -- \\
		0.6
		& 95.75 & 95.80 & 94.33 & -- & -- & -- \\
		0.8
		& 95.58 & 94.17 & -- & -- & -- & -- \\
		1.0
		& 93.80 & -- & -- & -- & -- & -- \\
		\bottomrule
	\end{tabular*}
\end{table}

Under the default setting, $(\lambda_{\mathrm{vec}},
\lambda_{\mathrm{intent}},\lambda_{\mathrm{sym}})=(8,7,20)$ corresponds
to the normalized ratio $(0.229,0.200,0.571)$. The results suggest that
the current setting lies in a relatively stable region on 2Wiki.

\subsection{Lexical Contribution Ablation}
\label{appendix:lexical_ablation}

The numerical weights in Eq.~\ref{eq:slocal} alone cannot determine the
actual contribution of each signal. We therefore compare \emph{Full}
with \emph{w/o symbolic term}
($\lambda_{\mathrm{sym}}=0$,
$\lambda_{\mathrm{vec}}=8$, and
$\lambda_{\mathrm{intent}}=7$)
and \emph{Symbolic-only} on the three validation sets. The results are
reported in Table~\ref{tab:lexical_ablation}.

\begin{table}[!htb]
	\centering
	\caption{Controlled lexical ablation on the three validation sets
		(Path Recall@5 / Path Recall@20).}
	\label{tab:lexical_ablation}
	\footnotesize
	\setlength{\tabcolsep}{1.5pt}
	\renewcommand{\arraystretch}{1.08}
	\begin{tabular*}{\columnwidth}
		{@{\extracolsep{\fill}}lccc@{}}
		\toprule
		\textbf{Dataset}
		& \textbf{Full}
		& \textbf{\shortstack{w/o symbolic\\term}}
		& \textbf{Symbolic-only} \\
		\midrule
		MuSiQue
		& 76.40 / 86.90
		& 74.80 / 86.60
		& 34.90 / 34.90 \\
		2Wiki
		& 89.05 / 95.80
		& 83.45 / 90.75
		& 85.18 / 85.28 \\
		HotpotQA
		& 86.65 / 96.60
		& 84.75 / 95.90
		& 53.70 / 53.70 \\
		\bottomrule
	\end{tabular*}
\end{table}

Removing the symbolic term reduces Path Recall on all three datasets,
showing that lexical calibration makes a practical contribution.
Symbolic-only also performs below Full on all three datasets, with
particularly large gaps on MuSiQue and HotpotQA. These results show that
lexical matching does not independently drive the full model's
performance. The full model benefits from the complementary contributions of
symbolic calibration, vector relevance, and residual-intent signals.

\subsection{Instruction-Template Controls}
\label{appendix:instruction_controls}

To examine whether the effect comes only from the presence of an
instruction-like prefix, we conduct experiments using different instruction templates while
fixing the other retrieval settings. \emph{Full} denotes the
task-oriented instruction used in IN-Retriever; \emph{T1}, \emph{T2},
and \emph{T3} denote generic, task-irrelevant, and wrong-direction
instructions, respectively. The four templates are as follows.

\begin{itemize}
	\setlength{\itemsep}{0.2em}
	\setlength{\topsep}{0.3em}
	\setlength{\parsep}{0pt}
	
	\item \textbf{Full task-oriented:}
	``Identify the key evidence in the context that helps find the next
	context to finally answer the question. Focus specifically on the
	entity type requested in the query.''
	
	\item \textbf{T1 Generic:}
	``Identify the relevant information in the context that helps
	understand the query before answering the question. Focus specifically
	on the content and wording provided in the context.''
	
	\item \textbf{T2 Task-irrelevant:}
	``Identify the stylistic information in the context that helps
	describe the writing before answering the question. Focus specifically
	on punctuation, capitalization, and sentence structure in the context.''
	
	\item \textbf{T3 Wrong-direction:}
	``Identify the final answer in the context that helps answer the
	question directly. Focus specifically on answer content rather than
	evidence needed to find the next context.''
\end{itemize}

The corresponding Path Recall@20 results are reported in
Table~\ref{tab:instruction_controls}.

\begin{table}[!htb]
	\centering
	\caption{Instruction-template controls on the three validation sets
		(Path Recall@20).}
	\label{tab:instruction_controls}
	\small
	\setlength{\tabcolsep}{2pt}
	\renewcommand{\arraystretch}{1.08}
	\begin{tabular*}{\columnwidth}
		{@{\extracolsep{\fill}}lcccc@{}}
		\toprule
		\textbf{Dataset}
		& \textbf{Full}
		& \textbf{T1}
		& \textbf{T2}
		& \textbf{T3} \\
		\midrule
		2Wiki
		& 95.80
		& 93.30
		& 83.23
		& 91.98 \\
		MuSiQue
		& 86.90
		& 86.85
		& 83.65
		& 86.80 \\
		HotpotQA
		& 96.60
		& 96.45
		& 94.55
		& 96.60 \\
		\bottomrule
	\end{tabular*}
\end{table}

T1's generic instruction lacks explicit guidance on next-hop evidence
and entity type. T2 redirects attention to textual style and therefore
produces the largest decrease. T3 focuses on the final answer rather than
the bridging evidence used to locate the next document, thereby weakening
next-hop retrieval guidance. These results show that the task-oriented
instruction is more consistent with the objective of our bridging clue
probing and further support its effectiveness.

\begin{table*}[!t]
	\caption{Ablation study of IN-Retriever on three multi-hop datasets in terms of Recall@5.}
	\label{tab:ablation_study}
	\centering
	\resizebox{\textwidth}{!}{%
		\begin{tabular}{l ccc ccc}
			\toprule
			& \multicolumn{3}{c}{\textbf{Components}} & \multicolumn{3}{c}{\textbf{Datasets (Recall@5)}} \\
			\cmidrule(lr){2-4} \cmidrule(lr){5-7}
			\textbf{Method} & \textbf{Steering} & \textbf{Residual} & \textbf{Cumulative} & \textbf{MuSiQue} & \textbf{2Wiki} & \textbf{HotpotQA} \\
			\midrule
			IN-Retriever (Full Model) & $\surd$ & $\surd$ & $\surd$ & \textbf{76.45} & \textbf{89.10} & \textbf{86.35} \\
			w/o Cumulative Scoring & $\surd$ & $\surd$ & -- & 75.05 & 82.95 & 84.20 \\
			w/o Residual-based Reranking & $\surd$ & -- & $\surd$ & 75.95 & 88.95 & 85.60 \\
			w/o Intrinsic Steering & -- & $\surd$ & $\surd$ & 66.75 & 71.10 & 77.95 \\
			\midrule
			Direct Retrieval (Baseline) & -- & -- & -- & 63.80 & 72.82 & 86.10 \\
			\bottomrule
		\end{tabular}
	}
\end{table*}

\subsection{Attention Patterns across Different Embedding Models}

\begin{figure*}[!ht]
  \centering

  \begin{subfigure}[t]{0.48\textwidth}
    \centering
    \includegraphics[width=\linewidth]{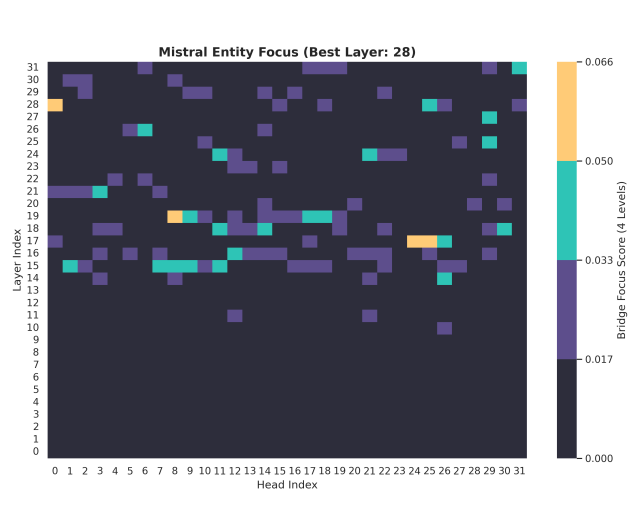}
    \caption{Mistral.}
    \label{fig:backbone_bfs_mistral}
  \end{subfigure}
  \hfill
  \begin{subfigure}[t]{0.48\textwidth}
    \centering
    \includegraphics[width=\linewidth]{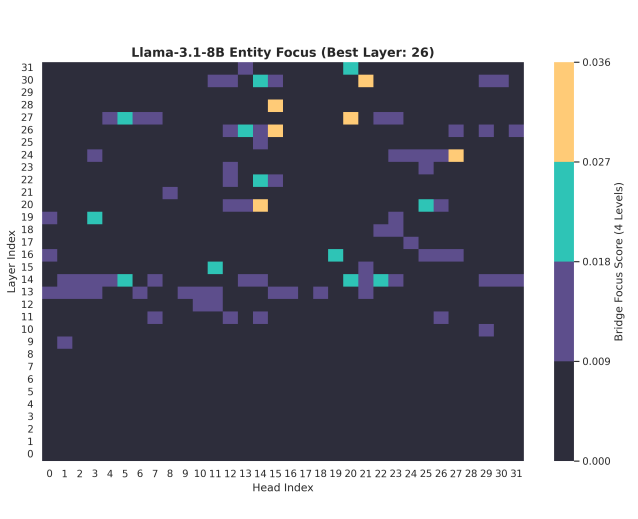}
    \caption{Llama.}
    \label{fig:backbone_bfs_llama}
  \end{subfigure}

  \caption{BFS heatmaps across all attention layers under two embedding backbones on the 2Wiki dataset.}
  \label{fig:backbone_bfs}
\end{figure*}

To examine whether the BFS distribution pattern is specific to a particular backbone model, we repeat the experiment described in Section \ref{subsubsec:layerWiseSensitivity} on two other mainstream embedding models, namely Mistral and Llama. We keep the same instruction template, bridging clue annotation protocol, BFS definition, and sampling procedure. Based on the resulting BFS values, we visualize the results in Figure~\ref{fig:backbone_bfs}.

From Figure~\ref{fig:backbone_bfs}, we find that, consistent with the results of Qwen3-Embedding-4B, attention heads with high BFS values are mainly concentrated in the middle-to-late layers of each model, while such heads are extremely sparse in the shallow layers. It suggests that the layer-wise sensitivity of bridging-clue-related attention signals is not unique to a single backbone, but instead reflects a broader pattern across architectures. It also indicates that internal signals relevant to bridging clues are more likely to emerge in layers where a certain degree of semantic integration has already taken place.

On the other hand, the figure shows that attention heads with high BFS values are not uniformly distributed across the entire parameter space, but are concentrated in only a small number of layers and heads. The observation is consistent with the sparse distribution of high-BFS heads in Qwen3-Embedding-4B, indicating that this pattern is unlikely to be an accidental property of a single model.

\subsection{Ablation Study}
\label{subsection:ablationStudy}

To assess the contributions of each component, we conduct an ablation
study by removing the three main modules one by one and report Recall@5
in Table~\ref{tab:ablation_study}.

After removing internal-attention-based clue extraction (w/o Intrinsic Steering), the fine-grained bridging clue $e$ is no longer available for residual calculation and cumulative scoring. To keep the remaining modules executable, we replace $e$ with a coarse document-level representation of the previous-hop document. In this setting, the model expands the evidence chain based on the entire previous document rather than on the specific clue localized by internal attention. As shown in Table~\ref{tab:ablation_study}, this variant leads to the largest performance drop across all datasets: Recall@5 decreases by 9.70, 18.00, and 8.40 percentage points on MuSiQue, 2Wiki, and HotpotQA, respectively. These results indicate that fine-grained bridging clues extracted from internal attention provide more targeted next-hop guidance than coarse document-level expansion, and constitute the primary source of the performance gains achieved by our method.  %

The cumulative scoring mechanism considers not only semantic relevance but also keyword matching. As shown in Table \ref{tab:ablation_study}, removing the cumulative scoring mechanism (w/o Cumulative Scoring) reduces Recall@5 by 1.40, 6.15, and 2.15 percentage points on the three datasets. The reductions suggest that relying solely on dense semantic similarity is insufficient to distinguish some key bridging entities in multi-hop retrieval. Explicit constraints introduced by keyword matching effectively complement this weakness and are helpful for the precise assessment of key entities and clues.

Finally, when removing the semantic residual step (w/o Residual Projection/Reranking), we disable the projection-based residual computation and remove its use in candidate reranking. In this variant, the model directly follows the documents retrieved by the bridging clue, without using the unresolved part of the original query as an additional constraint. As shown in Table~\ref{tab:ablation_study}, the Recall@5 of this variant consistently decreases across the three datasets. This suggests that the residual signal helps refine clue-based retrieval by reducing repeated focus on already covered evidence and by keeping the search aligned with the remaining information need of the original query. Therefore, although the bridging clue provides the main expansion direction, the residual step offers a useful complementary constraint for selecting more appropriate next-hop documents.  %

Overall, the ablation study verifies the indispensability of three core modules: internal-attention-based clue extraction, cumulative scoring mechanism, and semantic residual-based reranking. 

\end{document}